\documentclass[journal, twocolumn]{IEEEtran}
\usepackage{ulem}
\usepackage{amsfonts}
\usepackage{amssymb}
\usepackage{graphicx}
\usepackage{amsmath}
\usepackage{mathrsfs}
\usepackage{multirow}
\usepackage{graphicx}
\usepackage{color}
\usepackage{makecell}
\usepackage{algorithmicx, algorithm}
\usepackage{algpseudocode}
\usepackage{bm}
\usepackage{cite}
\usepackage[hyphens]{url}
\usepackage[caption=false,font=footnotesize,subrefformat=parens]{subfig}
\graphicspath{{figures/}}
\begin{document}


\title{Short-term Road Traffic Prediction based on Deep Cluster at Large-scale Networks}

\author{Lingyi~Han,~\IEEEmembership{Student~Member,~IEEE,} Kan~Zheng,~\IEEEmembership{Senior~Member,~IEEE,} Long~Zhao,~\IEEEmembership{Senior~Member,~IEEE,} Xianbin~Wang,~\IEEEmembership{Fellow,~IEEE}, 
    and Xuemin Shen,~\IEEEmembership{Fellow,~IEEE}
    \thanks{L.~Han, K.~Zheng, and L.~Zhao are with the Intelligent Computing and Communication (IC$^2$) Lab, Wireless Signal Processing and Networks Lab (WSPN) Key Lab of Universal Wireless Communications, Ministry of Education, Beijing University of Posts \& Telecommunications, Beijing, China, 100088. Contact email: $zkan@bupt.edu.cn$}
	\thanks{X. Wang is with Department of Electrical and Computer Engineering, University of Western Ontario, London, ON N6A 5B9, Canada.}
     \thanks{X. Shen is with Department of Electrical and Computer Engineering, University of Waterloo, Waterloo, ON N2L 3G1, Canada.}
	\thanks{Corresponding author: Kan~Zheng.}}

\maketitle
\begin{abstract}
Short-term road traffic prediction (STTP) is one of the most important modules in Intelligent Transportation Systems (ITS). However, network-level STTP still remains challenging due to the difficulties both in modeling the diverse traffic patterns and tacking high-dimensional time series with low latency. Therefore, a framework combining with a deep clustering (DeepCluster) module is developed for STTP at large-scale networks in this paper. The DeepCluster module is proposed to supervise the representation learning in a visualized way from the large unlabeled dataset. More specifically, to fully exploit the traffic periodicity, the raw series is first split into a number of sub-series for triplets generation. The convolutional neural networks (CNNs) with triplet loss are utilized to extract the features of shape by transferring the series into visual images. The shape-based representations are then used for road segments clustering. Thereafter, motivated by the fact that the road segments in a group have similar patterns, a model sharing strategy is further proposed to build recurrent NNs (RNNs)-based predictions through a group-based model (GM), instead of individual-based model (IM) in which one model are built for one road exclusively. Our framework can not only significantly reduce the number of models and cost, but also increase the number of training data and the diversity of samples. In the end, we evaluate the proposed framework over the network of Liuli Bridge in Beijing. Experimental results show that the DeepCluster can effectively cluster the road segments and GM can achieve comparable performance against the IM with less number of models.
\end{abstract}


\section{Introduction}
\label{sec_intro}
The short-term road traffic prediction (STTP) technique has been studied in achieving efficient route planning and traffic control in Intelligent Transportation Systems (ITS) recently~\cite{rg2}. The main idea of STTP is to predict the road traffic state (i.e., flow, speed and density) in the next five to thirty minutes by analyzing historical data~\cite{sttp}. However, existing STTP studies mainly focused on one road segment, or a small-scale network containing several adjacent road segments, which is opposite to the effective route planning that requires a global perspective based on the information of the whole network~\cite{review1, network-w1, network-w2}. Besides, the majority of existing STTP algorithms are limited to a single scenario such as freeway, arterial or corridor, which are difficult to be generalized to a heterogeneous road network. The past STTP method for large-scale road network is to develop a specific model for each road segment termed as individual-based model (IM), or a general model for all road segments termed as whole-based model (WI). Since the multiplicity and heterogeneity of the large-scale network, neither of the two models is appropriate for the large-scale networks. Firstly, too many IMs will take up lots of storage resources in ITS. Secondly, a WI is not competent for modeling the whole network with different types of traffic patterns. Moreover, the development of ITS over the city increases the number of traffic data in terms of time span and granularity~\cite{bigdata}. Making full use of the big traffic data to improve the performance of the prediction becomes a challenge. Therefore, a feasible STTP at large-scale network needs to be studied.

\par
Generally, representation learning, a.k.a. dimension reduction, is used to transform the raw data into a good representation that makes the subsequent tasks easy. It plays an important role in time series clustering, because time series are essentially high-dimensional and susceptible to noise. Hence, clustering directly with raw series is computationally expensive and distance measures are highly sensitive to the distortions. Recently, deep learning (DL) has been developed with great success in many areas, including computer vision, speech recognition and natural language processing due to its theoretical function approximation properties~\cite{approximation} and demonstrated feature learning capabilities~\cite{representation}. Therefore, deep representation is used for traffic series clustering.
\par
In this paper, a feasible framework composed of a deep clustering module and several prediction models is proposed for STTP at large-scale networks. More specifically, a shape-based representation learning method is developed for road segments clustering. On the other hand, several predictions are combined to achieve the STTP at the network. The main contributions of the paper are summarized as follows:


\begin{itemize}
\item By fully exploiting the periodicity of traffic patterns, we propose a method to generate triplets from unlabeled dataset. The raw traffic series are divided into sub-series by periods, three of which are selected to generate a triplet according to a specific criterion. The dimension of sub-series used for representation learning is significantly reduced, compared to raw series.
\item A supervised deep clustering module termed as DeepCluster, is developed. Unlike the existing hand-craft features, such as the frequency transformation, wavelet transformation, Shapelets \textit{et al.}, a pure data driven method is proposed to learn the shape-based representations of traffic series in a visualized way. A rasterization strategy is first designed to transform the traffic series into traffic images. A convolutional neural network (CNN) with triplet loss is then used for representation learning. At last, the representations are used to cluster the network into $K$ groups by traditional clustering methods.\\
\item Based on the idea of model sharing, $K$ group-based models (GMs) that are constituting a prediction at network is proposed to achieve a good tradeoff between the quantity of models and the performance of predictions. Specifically, all road segments in one group share one prediction model and each GM allows the training samples generated by the road segment from the same group to be aggregated to learn the model. Model sharing increases the number and the diversity of the training samples, which is beneficial for DNNs training. The experiment results validate that the GM has stronger generalization ability than IM. We also analyze the impact of input interval on performance by experiments.\\
\end{itemize}
The rest of paper is organized as follows. Section~\ref{sec:related} reviews the related works. In Section~\ref{sec:data}, the data used throughout the paper is  described. Section~\ref{sec:system} formulates the STTP problem at large-scale network. In Section~\ref{sec:method}, the DL methodologies are introduced. The proposed framework of STTP including DeepCluster and DeepPrediction is then proposed in Section~\ref{sec:framework}. In Section~\ref{sec:evaluation}, simulation results demonstrating the performance of the proposed framework are given, before concluding the paper in Section~\ref{sec:conclusion}.

\section{Related Works}\label{sec:related}

\subsection{Time Series Representation Learning}\label{sec:rl}
A wide \text{color}{variety} of methods had been developed for time series representation learning in clustering~\cite{r1, r2,r3}, such as spectral transformation~\cite{st}, wavelets transformation~\cite{st}, eigenvalue analysis techniques~\cite{svd}, piecewise linear approximation (PLA)~\cite{pla}, adaptive piecewise constant approximation (APCA)~\cite{apca}, symbolic approximation (SAX)~\cite{sax}, piecewise aggregate approximation (PAA)~\cite{paa}, perceptually important point (PIP)~\cite{pip} \textit{et al}. However, all these methods are hand-craft features, which are  designed to describe specific time series pattern and heavily rely on the database. 
\par
A new trend appears with artificial neural networks (ANNs), especially deep NNs (DNNs) based representation learning in clustering, which are data-driven and capable of learning a powerful representation from raw data through a high-level and non-linear mapping. Therefore, some works have used the deep representation learning to improve clustering performance. C. Song \textit{et al}. in~\cite{dl1} integrated $K$-means algorithm into a stacked auto-encoder (SAE) by minimizing the reconstruction error as well as the distance between data points and corresponding clusters. It alternatively learned the representations and updated cluster centers. In~\cite{dl2,dl3}, the $k$-means algorithm used the nonlinear representations that are learned by DNNs for clustering. J. Xie \textit{et al}. in~\cite{dl4} proposed a deep embedded clustering that simultaneously learned the representations and cluster  assignments by defining a centroid-based probability distribution and minimizing its Kullback-Leibler (KL) divergence to an auxiliary target distribution. K. Tian \textit{et al}. in~\cite{dl5} improved the existing works by proposing a general flexible framework that integrated traditional clustering methods into different DNNs. The framework is optimized by alternating direction of multiplier method (ADMM). 
However, the above methods all worked with the static data that is simple and low dimensional compared with time series data in general. On the other hand, there is less research on the deep representation learning of time series in clustering. Therefore, an efficient time series representation learning algorithm dedicated for clustering needs to be developed.

\subsection{Short Term Traffic Prediction}
\label{sec:tp}
There are numerous researches on single-point STTP~\cite{review1}, such as autoregressive integrated moving average (ARIMA) family of models, regression models, Markov models, Kalman filters, Bayesian networks, traffic flow theory-based simulation models and ANNs. 
Obviously, single-point models predict the future traffic state for a target road segment only using its own historical data, which ignores the relations between the target road segment and adjacent segments. Consequently, some researches have focused on predicting one or multiple segments by taking the spatio-temporal interrelations between adjacent road segments into account~\cite{lasso, st1, st2, DCRNN}. However, the above network-level STTP researches are restricted to small regions that containing several adjacent road segments.
\par
Recently, a few literatures begin to pay attention to the predictions at the large-scale networks. In~\cite{simulation1, simulation2}, dynamic simulator based on traffic flow theory was used for STTP at the whole network with limited traffic data. ~\cite{compression1, compression2, compression3, compression4} only predicted the traffic state of the representative road subset to achieve the prediction at the whole network by utilizing data compression technologies. However, the performance of prediction was poor resulted from compression and reconstruction errors. Min \textit{et al}. in~\cite{l500} considered a road network consists of about 500 road segments. However, they developed a custom model for the test area, which is not practical. M. Asif \textit{et al}. in \cite{svr5000} performed prediction for each individual road segment with support vector regression (SVR) algorithm over a large network containing 5,000 road segments. Then $K$-means algorithm was used to cluster the road segments to analyze the spatial prediction performance. But the prediction method may not work well, since the performances differed greatly among clusters and the mean error of one cluster is up to $17.18\%$ of five-minute prediction. Besides, STTP for each individual road segment is hard to implement on large-scale networks in practice. X. Ma \textit{et al}. in~\cite{images} proposed a CNN-based method that arranges the traffic data into 2D (2-dimensional) matrices as inputs to predict the large-scale traffic speeds. However, they only built one model and expected it to fit for all segments without considering the fact that the whole network is heterogeneous with different type of segments. Therefore, these attempts are hard to be implemented on large-scale networks with high accuracy. 

\begin{figure}[t]
	\centering\includegraphics[width=3.5in]{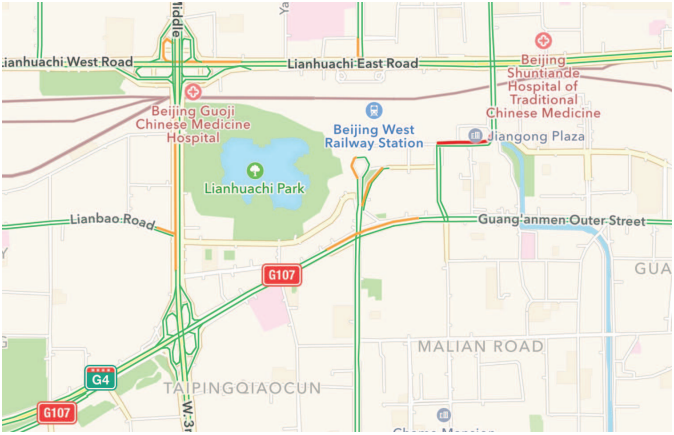}
	\caption{The topology of the network at Liuli Bridge, Beijing.}\label{fig:map}
\end{figure}

\begin{figure}[!t]
	\centering\includegraphics[width=3.5in]{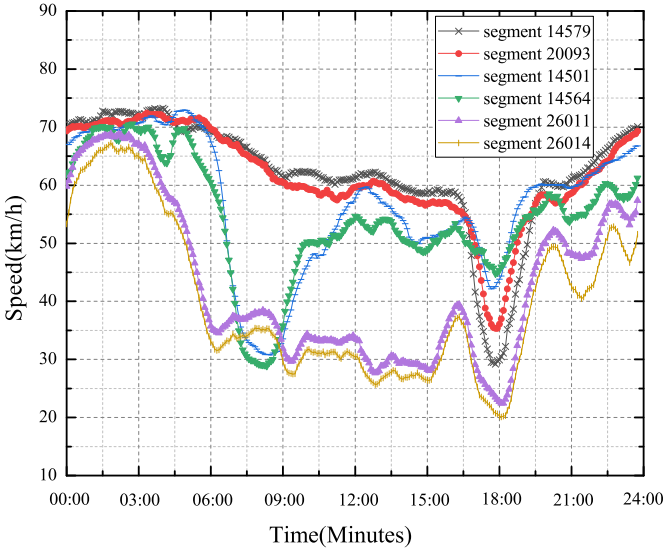}
	\caption{The every five-minute average traffic speeds of six road segments from September, 2017 to November, 2017.}\label{fig:heterogeneity}
\end{figure}

\section{The Data}\label{sec:data}
The traffic data used throughout the paper is described in this section. The topology of Liuli Bridge is shown in Fig.~\ref{fig:map}. The network consists of about 1,000 road segments with a diverse level of road functions including express way, arterial road, access road, side road \textit{et al}. In addition, the dataset collected by Beijing Transportation Institute contains the traffic speed data from September, 2017 to November, 2017 with five-minute sampling interval. Hence, it has totally $90 \times 288$ measured data, where $90$ means the total number of days and $288$ means the number of values collected in each day. The data is measured by vehicles that are equipped with GPS such as taxis and buses.

\section{Formulation of STTP Problem}\label{sec:system}
Consider a large-scale network $\Phi$ consisting of $N_{\text{r}}$ road segments, i.e., $\Phi = \{ \bm{x}^{(r)} \}_{r=1}^{N_{\text{r}} }$, where $\bm{x}^{(r)} = [x_{1}^{(r)}, x_{2}^{(r)},\dots,x_{N_{\text{t}}}^{(r)}]$ is a time series of $N_\text{t}$ measurements at segment $r$. We denote a sub traffic series by
\begin{equation}\label{e:sub}
\bm{x}_{t: L:l} = [x_{t}, x_{t+l},\dots, x_{t+(L-1)l}],
\end{equation} 
where $\bm{x}_{t : L:l}$ is a set of $L$ continuous measured values with intervals $l$ from a time series $\bm{x}$, that starts at position $t$ with $1 \leq t \leq N_\text{t}$, $1\leq l \leq N_\text{t}$ and $1\leq L \leq N_\text{t}$. $\bm{x}_{t :L :1}$ is abbreviated as $\bm{x}_{t:L}$ for simplicity.
\par
Let $\hat{x}_{t+N_{\text{o}}}$ be the forecast of traffic state of the prediction horizon $N_{\text{o}}$, given the corresponding $N_{\text{i}}$ historical measurements up to time $t$. The goal of STTP is to construct a mapping $f(\cdot)$ between the historical traffic state and the future one, i.e., \begin{equation}\label{e:stl}
	\begin{aligned}
\hat{x}_{t+N_{\text{o}}} &= f(x_{t-N_{\text{i}}}, x_{t-N_{\text{i}}+1},\dots, x_{t})\\
 &= f(\bm{x}_{t-N_{\text{i}}: N_{\text{i}}}).
	\end{aligned}
\end{equation} 
\par
As stated above, IM and WM are both inappropriate for the large-scale networks, because they not only consist of a large number of road segments, but also a variety of types of road segments as shown in Fig.~\ref{fig:heterogeneity}. On one hand, it's unpractical to construct and store massive amounts of IMs in ITS. Besides, the number of training samples collected from one segment is insufficient to learn a robust DL model. On the other hand, it's impossible to build a model for the whole network with different types of traffic pattens. In addition, the model is vulnerable to the curse of dimensionality by taking historical data from all segments as inputs. Then how to make a proper utilization of the tremendous traffic data to achieve the effective and practical STTP is still a problem.
\par
To tackle this problem, we cluster the road segments into groups, each of which has a typical traffic pattern. Within each group, the traffic patterns of all road segments are highly similar in shape. Based on that, a STTP model is built for a group, rather than a segment or whole network. The challenges in our problem include \textit{i}) representation learning of the traffic series that are high-dimensional and sensitive to distortion, and \textit{ii}) representation learning from unlabeled traffic data that are beneficial to cluster task.

\begin{figure}[t]
	\centering\includegraphics[width=3.5in]{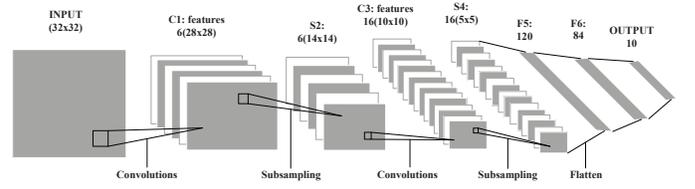}
	\caption{Architecture of LeNet-5. Above the rectangles are the number of channels and its size in parenthesis.}\label{fig:lenet}
\end{figure}

\section{Deep Learning for Traffic Predictions}\label{sec:method}
In this section, we deals with the tremendous traffic series by means of the DL technologies, including CNNs and recurrent NNs (RNNs), which will be explained in this section.

\subsection{Convolutional Neural Networks}
The key aspect of CNNs is that the features are not designed by human engineers, but are learned from data using a general-purpose learning procedure~\cite{representation}. Fig.~\ref{fig:lenet} shows the architecture of a typical CNN, named LeNet-5. CNNs can take any form of arrays, such as 1D series, 2D images and 3D videos as inputs. A CNN is made up of layers, where two main types of layers different with the regular ANNs are convolutional layers (C layers in Fig.~\ref{fig:lenet}) and subsampling layers (S layers in Fig.~\ref{fig:lenet}).
\par
In the $l$-th convolutional layer $\text{C}^{(l)}$, the outputs of the previous layer are fed to convolve with several convolutional kernels. After that, the outputs are added by biases and activated by a nonlinear function to form new representations (features in Fig.~\ref{fig:lenet}) for the next layer. Assuming the current layer accept an input volume $\bm{O}^{(l-1)}$ of size $W_{\text{o}}^{(l-1)} \times H_{\text{o}}^{(l-1)} \times D_{\text{o}}^{(l-1)}$. Formally, the output $\bm{O}^{(k,l)}$ of size $W_{\text{o}}^{(l)} \times H_{\text{o}}^{(l)} \times 1$ filtered by the $k$-th kernel $\bm{K}^{(k,l)}$ of size $W_{\text{k}}^{(l)} \times H_{\text{k}}^{(l)} \times 1$ with stride $s$ is given by
\begin{equation}\label{e:cnn}
\bm{O}^{(k,l)} = g(\bm{K}^{(k,l)} \otimes \bm{O}^{(l-1)}+b^{(k,l)}), k=1,2,\dots,N_{\text{k}}^{(l)}
\end{equation}
where $N_{\text{k}}^{(l)}$ is the number of kernels and $b^{(k,l)}$ is a bias of layer $\text{C}^{(l)}$, respectively. $\otimes$ represents a discrete convolution operator . $g(\cdot)$ is a activation function such as tanh functuon, relu function \textit{et al}. By concatenating $\bm{O}^{(k, l)}$ along the last dimension, the output $\bm{O}^{(l)}$ for layer $\text{C}^{(l)}$ of the size $W_{\text{o}}^{(l)} \times H_{\text{o}}^{(l)} \times D_{\text{o}}^{(l)}$ can be derived, and both of which can be calculated by
\begin{equation}\label{e:cw}
W_{\text{o}}^{(l)}= \lfloor \frac{W_{\text{o}}^{(l-1)} - W_{\text{k}}^{(l)}}{s}\rfloor+1,
\end{equation} 
\begin{equation}\label{e:ch}
H_{\text{o}}^{(l)} = \lfloor \frac{H_{\text{o}}^{(l-1)}-H_{\text{k}}^{(l)}}{s} \rfloor+1,
\end{equation} 
\begin{equation}\label{e:cd}
D_{\text{o}}^{(l)}=N_{\text{k}}^{(l)},
\end{equation} 
where $\lfloor \cdot \rfloor$ represents rounded down. With parameter sharing, there are $N_{\text{w}}^{(l)}$ learnable weights of layer $C^{(l)}$ in total,
\begin{equation}\label{e:weight}
N_{\text{w}}^{(l)} = W_{\text{k}}^{(l)} \times H_{\text{k}}^{(l)} \times D_{\text{o}}^{(l-1)} \times N_{\text{k}}^{(l)} +  N_{\text{k}}^{(l)}.
\end{equation}
\par
In the $(l+1)$-th subsampling layer $\text{S}^{(l+1)}$, the spatial resolution of representations is reduced to increase the level of distortion invariance. After layer $\text{C}^{(l)}$, the layer $S^{(l+1)}$ accepts a volume of size $W_{\text{o}}^{(l)} \times H_{\text{o}}^{(l)} \times D_{\text{o}}^{(l)}$ as input. Specifically, representations in the previous layer are pooled over neighborhood within a rectangular region of $W_{\text{s}}^{(l+1)} \times H_{\text{s}}^{(l+1)}$, by either a max-pooling function 
\begin{equation}\label{e:maxp}
\begin{aligned}
O_{i,j,k}^{(l+1)} = \max_{\substack {i \le p \le i+W_{\text{s}}^{(l+1)} \\ j\le q \le j+H_{\text{s}}^{(l+1)}}} 
(O_{p,q,k}^{(l)}), \\ i \le W_{\text{o}}^{(l+1)}, j \le H_{\text{o}}^{(l+1)}, k \le D_{\text{o}}^{(l+1)}
\end{aligned}
\end{equation} 
or an average-pooling function. 
where $\bm{O}^{(l+1)}$ is the output of size $W_{\text{o}}^{(l+1)} \times H_{\text{o}}^{(l+1)} \times D_{\text{o}}^{(l+1)} $, and both of them can be calculated by
\begin{equation}\label{e:sw}
W_{\text{o}}^{(l+1)}= \lfloor \frac{W_{\text{o}}^{(l)} - W_{\text{s}}^{(l+1)}}{s} \rfloor+1,
\end{equation} 
\begin{equation}\label{e:sh}
H_{\text{o}}^{(l+1)} = \lfloor \frac{H_{\text{o}}^{(l)}-H_{\text{s}}^{(l+1)}}{s} \rfloor+1,
\end{equation} 
\begin{equation}\label{e:sd}
D_{\text{o}}^{(l+1)} = D_{\text{o}}^{(l)}.
\end{equation}
\par
The convolutional and subsampling operators make the new representations more invariance to the distortion compared to the raw data. Besides, the parameter sharing make the CNNs capable of processing high-dimensional inputs. The aforementioned characteristics allow to adopt the CNNs for time series representation learning.In this section, we explore an efficient deep CNN architecture, FaceNet~\cite{facenet} to learn the deep representations of the raw time series.

\begin{figure}[t]
	\centering\includegraphics[width=3.5in]{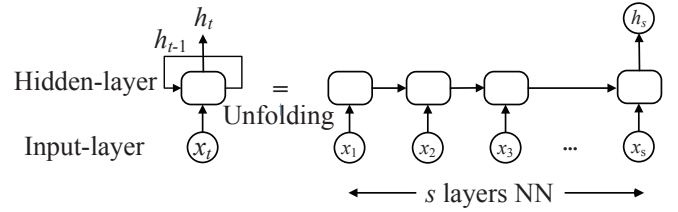}
	\caption{Architecture of a basic three-layer RNN.}\label{fig:rnn}
\end{figure}

\begin{figure*}[!t]
	\centering\includegraphics[width=6in]{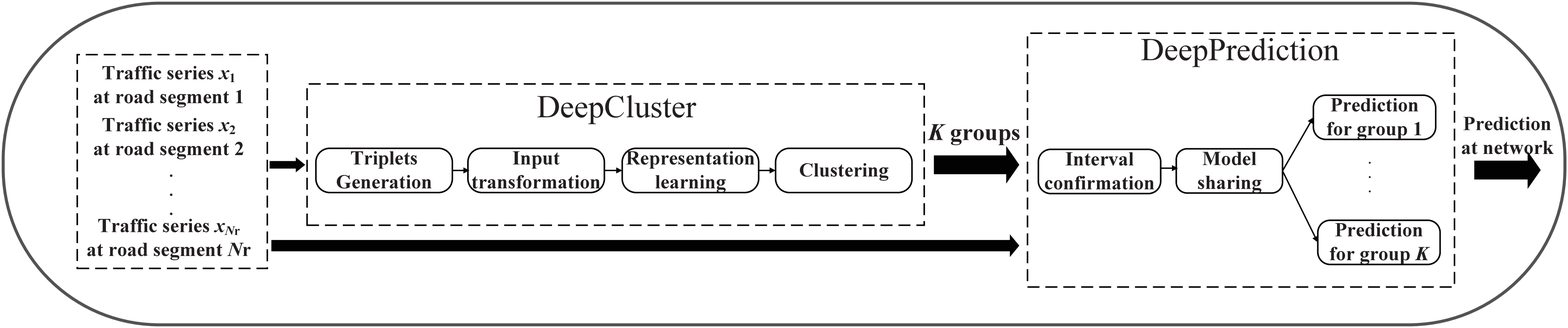}
	\caption{The block diagram of representation learning.}\label{fig:overview}
\end{figure*}

\subsection{Recurrent Neural Networks}
Unlike the regular ANNs, RNNs are capable of exhibiting the temporal correlations of time series, which makes them applicable to tasks such as language modeling, speech recognition or time series forecasting.
\par
Assuming the duration of the temporal correlations (defined as time step) is $s$, a three-layer RNN can be regarded as a $s$-layer feed-forward NN by unfolding it through time, As shown in Fig.~\ref{fig:rnn}. The RNN reads a series $\bm{x}_{1:s}$ one by one and each RNN block takes a value at one time as input. The current hidden state $\bm{h}_{t}$ at time $t$ is computed from the current input $x_{t}$ and the previous hidden state $\bm{h}_{t-1}$ by
\begin{equation}\label{e:sd}
\begin{aligned}
\bm{h}_{t} &= g( x_{t}, \bm{h}_{t-1})\\
& = g(x_{t}, g(x_{t-1}, \bm{h}_{t-2}))\\
& = \cdots
\end{aligned}
\end{equation} 
where $\bm{h}_{t-2}$ is the hidden state of the last two RNN blocks. $g(\cdot)$ is the activation function of the hidden layer. The key idea of RNNs is to imitate a sequential dynamic behavior with a chain-like structure that allows the information to be passed from previous layer to the current one. In this paper, RNNs are used to model the temporal correlations of traffic series.

\section{Proposed Framework for STTP}\label{sec:framework}
In this section, a framework dedicated for STTP at large-scale networks is described in details. The architecture of this framework is shown in Fig.~\ref{fig:overview}. It consists of two major components, i.e., DeepCluster and DeepPrediction. The inputs are historical traffic states with fixed interval coming from different road segments, while the outputs are predictions for a given time period. The inputs are fed to the DeepCluster module, and are divided into several groups. Afterwards, the DeepPrediction module performs the predictions for the network. 

\begin{figure}[!t]
	\centering\includegraphics[width=3.5in]{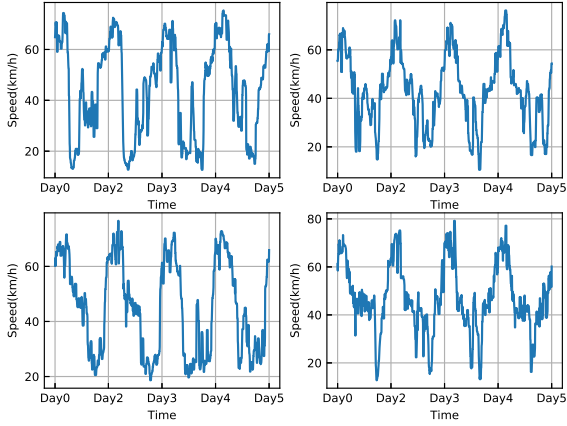}
	\caption{The every five-minute average traffic speeds of four road segments on different days of the week from September, 2017 to November, 2017.}\label{fig:recurrent}
\end{figure}

\begin{figure}[!t]
	\centering\includegraphics[width=3in]{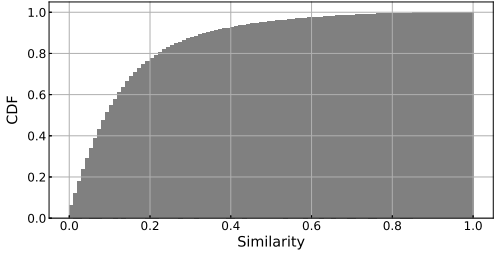}
	\caption{The CDF of similarity with real traffic speed.}\label{fig:cdf}
\end{figure}

\subsection{DeepCluster}\label{sec:dc}
Traffic series clustering method at large-scale networks is first proposed, which is implemented via deep representation learning. Before developing the clustering algorithm, the problem of clustering at large-scale networks is formally defined as follows:
\par
Definition 1: Given a large-scale network $\Phi$ consists of $N_\text{r}$ traffic series, i.e., $\Phi = \{ \bm{x}^{(r)} \}_{r=1}^{N_{\text{r}} }$ the process of partitioning of $\Phi$ into $K$ groups $\{\bm{C}^{(1)}, \bm{C}^{(2)},\dots, \bm{C}^{(K)}\}$, is called \textit{traffic series clustering}. In such a way that homogenous traffic series are grouped together based on a certain similarity measure.
\par
\begin{figure}[!t]
	\centering\includegraphics[width=3in]{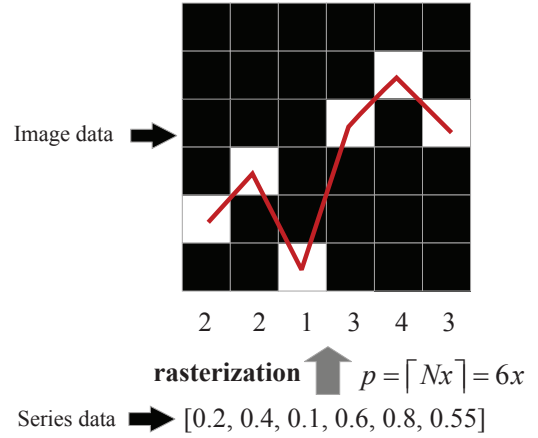}
	\caption{The schematic diagram of the inputs transformation.}\label{fig:rasterization}
\end{figure}

In contrast to the traditional extrinsic hand-craft features, human brains can seize the intrinsic visual-based features easily, which is why they can quickly distinguish different types of the time series under the help of high abstraction ability. Moreover, compared with raw time series, the intrinsic visual-based features are more steady. They are less affected by the distortions and the scale of samples. To address the issues of the raw data-based or hand-craft-based clustering methods, we use the deep representation learning for series clustering. The DNN is employed to learn a mapping from the raw high-dimensional traffic series to the low-dimensional representations that are used for clustering. 
\par
\begin{figure}[!t]
	\centering\includegraphics[width=3.5in]{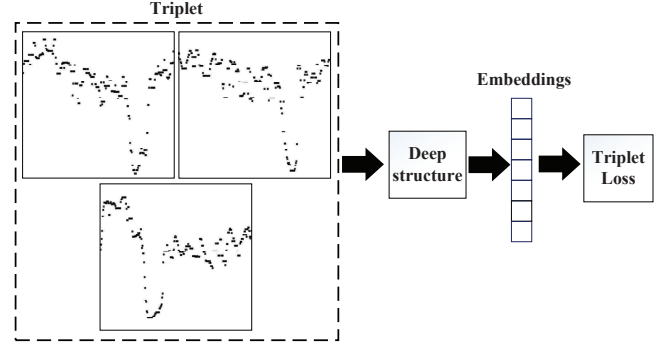}
	\caption{The block diagram of representation learning.}\label{fig:structure}
\end{figure}

The DeepCluster module includes triplets generation, inputs transformation, representation learning and clustering. Details of each step are given below.
\begin{enumerate}
	\item \textbf{Triplets Generation.} As can be seen in Fig.~\ref{fig:recurrent}, the traffic patterns follow the same trend among days. In order to study the traffic periodic pattern in a day, we calculate the traffic similarity defined in\cite{cdf}. The traffic similarity is defined as the normalized gaps between each pair of measurements in two consecutive days from one road segment. As stated in Sec.~\ref{sec:system}, traffic speeds are collected every 5 minutes. Since one day has 288 time intervals, the traffic similarity $\text{SIM}_{t^{*}}^{(r)}$ at segment $r$ in time slot ${t^{*}}$ can be calculated by
	\begin{equation}\label{e:similar}
	\text{SIM}_{t^{*}}^{(r)} = \frac{|x^{(r)}_{t^{*}}-x^{(r)}_{{t^{*}}+288}|}{\max_{\substack {1 \le t \le N_{\text{t}}-288}} |x^{(r)}_t-x^{(r)}_{t+288}|}.
	\end{equation} 
	The cumulative distribution function (CDF) of $\text{SIM}_{t^{*}}^{(r)}$ is shown in Fig.~\ref{fig:cdf}. We can see that more than $80\%$ $\text{SIM}_t^{(r)}$ are smaller than $0.2$, which indicates
	that periodic pattern exists in traffic series at most read segments.
    \par
	To fully exploit the traffic temporal features and periodic patterns, we split the traffic series into sub-series by periods, and generate triplets for representation learning. Given $N_{\text{r}}$ traffic series with period $N_{\text{p}}$ measured from $N_{\text{r}}$ road segments, we split the series into sub-series by periods, termed as periodic sub-series. Thus, we have $d=N_{\text{t}}/N_{\text{p}}$ periodic sub-series for each segment,
	\begin{equation}\label{e:split}
	 \bm{x}_{1:N_\text{p}}^{(r)}, \bm{x}_{N_\text{p}+1: N_\text{p}}^{(r)},\dots,\bm{x}_{(d-1)N_\text{p}+1: N_\text{p}}^{(r)},
	\end{equation} 
	here $\bm{x}_{jN_\text{p}+1:N_\text{p}}^{(r)} = [ x_{jN_\text{p}+1}^{(r)}, x_{jN_\text{p}+2}^{(r)},\dots,x_{(j+1)N_\text{p}}^{(r)} ]$ is the $(j+1)$-th periodic sub-series at segment $r$ with $\ 0 \le j  \le d-1$. A triplet is made up by randomly choosing two different periodic sub-series from one segment, and one sub-series from another segment,
   	\begin{equation}\label{e:triplet}
   	\begin{aligned}
   \{\bm{x}_{iN_\text{p}+1:N_\text{p}}^{(r_i)}, \bm{x}_{jN_\text{p}+1:N_\text{p}}^{(r_i)}, \bm{x}_{kN_\text{p}+1:N_\text{p}}^{(r_j)}\}. \\
   0 \le i, j, k \le d-1, i \not=j, r_i\not= r_j
   \end{aligned}
   \end{equation}
	\item \textbf{Inputs transformation.} In order to extract the features of shape, a rasterization strategy is designed to visualize the series into images shown in Fig.~\ref{fig:rasterization}. The transformed images can reveal the shape information of series well, such as bulge, sink and so on. Let the series $\bm{x} = [x_1, x_2,\dots,x_N]$ be standardized by min-max normalization to keep values between $0$ and $1$. A series is transformed to a matrix by expanding each element to a vector. For the $i$-th element $x_i$, the position $p_i$ at the $i$-th column of the matrix is,
	\begin{equation}\label{e:position}
        p_i = \lceil N x_i \rceil. \ \ \ p_i \in  \{1,2,\dots,N\}
	\end{equation} 
        The matrix $\bm{X}_{N, N}$ corresponding to the series $\bm{x}$ can be written as:  
	\begin{equation}\label{e:image}
    \bm{X}_{N, N} = [255_{N}(p_1), 255_{N}(p_2),\dots, 255_{N}(p_N)],
	\end{equation} 
	where $255_{N}(p_i)$ is a $N$-dimensional vector with the pixel value of $255$ at its $i$-th entry standing for white and $0$ standing for black elsewhere. The transformed image is shown in Fig.~\ref{fig:structure}. The matrixes are used as the inputs to the representation learning. The sub-image corresponding to the sub-series $\bm{x}_{iN_\text{p}+1:N_\text{p}}^{(r)}$ is represented as $\bm{X}_{i}^{(r)}$. Therefore, the triplet becomes,
	   	\begin{equation}\label{e:newtriplet}
   	\begin{aligned}
   \{\bm{X}_{i}^{(r_i)}, &\bm{X}_{j}^{(r_i)}, \bm{X}_{k}^{(r_j)}\}. \\
   0 \le i, j, k \le d&-1, i \not=j, r_i\not= r_j
   \end{aligned}
   \end{equation}	
	\item \textbf{Representation learning and clustering.} DNNs with triplet loss from~\cite{facenet} is employed to strive for a representations over a triplet, from an image space into a feature space. The triplet loss encourages the representations of a pair of sub-images from one segment to be close to each other in the feature space, and the those from different segments to be far away. The representation of $x$ is denoted by $f(x)$. Thus, the triplet loss that is being minimized is,   
   \begin{equation}\label{e:triloss}
   \begin{aligned}
   ||f(\bm{X}_{i}^{(r_i)})-f(\bm{X}_{j}^{(r_i)})||_{2}^{2}&-||f(\bm{X}_{i}^{(r_i)})-f(\bm{X}_{k}^{(r_j)})||_{2}^{2}, \\
   \forall \{\bm{X}_{i}^{(r_i)}, &\bm{X}_{j}^{(r_i)}, \bm{X}_{k}^{(r_j)}\} \in \Gamma
   \end{aligned}
   \end{equation}
where $\Gamma$ is the set of all possible triplets. The structure of DNNs with triplet loss is shown in Fig.~\ref{fig:structure}, where the outputs of the last layer are the representations used for clustering. The dimension of the representations in clustering is lower than the raw series. For example, considering a traffic series with five-minute interval during $90$ days. The length of whole series is $288\times 90$, while the length of daily sub-series is $288$. If we use $32$-dimensional representations in clustering, the ratio of reduction in dimension is about $\frac{32}{288\times 90} \approx 0.1\%$. Subsequently, we average all the representations from one road segment, and cluster the representations into $K$ groups, where $K$ is much less than $N_\text{r}$. Therefore, $N_{\text{r}}$ road segments are clustered into $K$ groups,
   \begin{equation}\label{e:group}
   \begin{aligned}
   \bm{C}^{(k)} = \{\bm{x}^{(r, k)}\}, 1\le k \le K, 1\le r \le N_\text{r}
   \end{aligned}
   \end{equation}
where $\bm{C}^{(k)}$ denotes the $k$-th group. $\bm{x}^{(r,k)}$ represents the $r$-th road segment in network $\phi$, which is clustered into group $\bm{C}^{(k)}$.
\end{enumerate}


\begin{figure}[!t]
	\centering\includegraphics[width=3in]{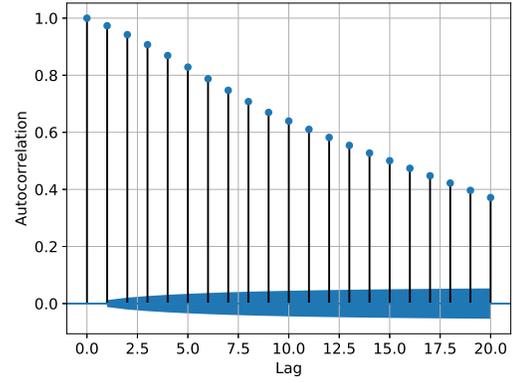}
	\caption{The ACF of the traffic speeds of a random segment from lag $1$ to lag $20$. The shaded area represents the 95$\%$ confidence intervals, which is used to determine whether the autocorrelation coefficients is significantly different from zero.}\label{fig:acf}
\end{figure}

\subsection{DeepPrediction}\label{sec:dp}
After partitioning the network into $K$ groups, we build a prediction model for a group in the DeepPrediction module. Some definitions and statements are first given.
\par
Definition 2: Given two functions $g^{(1)}: \mathbb{R} \to \mathbb{R}$ and $g^{(2)}: \mathbb{R} \to \mathbb{R} $, if $\hat{g}^{(1)}$ coincides with $g^{(2)}$ within a specified measurement range after horizontal translation, $g^{(1)}$ is \textit{homogeneous} with $g^{(2)}$.
\par
Statement 1: Given two \textit{homogeneous} functions $g^{(1)}: \mathbb{R} \to \mathbb{R}$ and $g^{(2)}: \mathbb{R} \to \mathbb{R}$, for simplicity assuming $g^{(1)}$ has coincided with $g^{(2)}$, and $N$ distinct successive samples $(x_{i},y_{i}^{(1)})\in \mathbb{R} \times \mathbb{R}$ generated from $g^{(1)}$. Construct a mapping between historical $y$ values and the future $y$ value: $f^{(1)}: [y_1^{(1)},y_2^{(1)},\dots,y_{N-1}^{(1)}]  \to y_{N}^{(1)}$. Similarly get $N$ successive  samples from $g^{(2)}$ at same $x$ values and construct the mapping $f^{(2)}: [y_1^{(2)},y_2^{(2)},\dots,y_{N-1}^{(2)}]  \to y_{N}^2$. It is obvious that $f^{(2)}$ is equal to $f^{(1)}$.
\par
Based on the Statement 1, we propose an idea of model sharing that all road segments within a group can share a prediction model. The implementation of the DeepPrediction is elaborated as follows:
\begin{enumerate}
	\item \textbf{Interval confirmation.} According to the periodicity, it is intuitive to use the measurements in a period to predict the next traffic state. In order to measure the autocorrelation between current and past traffic values, we calculate the autocorrelation function (ACF) at lag $i$, which is the correlation between series values that are $i$ intervals apart. As shown in Fig.~\ref{fig:acf}, the measurements are linearly correlated with the contiguous measurements. The high autocorrelations imply that importing all measurements in a period will result in information redundancy. We calculate the input interval $l$ by   
	 \begin{equation}
	 l =\max \limits_{p_{i}>p, i \ge 1 }\{i\},         
	 \end{equation}
	where $p_i$ denotes the ACF at lag $i$, and $p$ is the given threshold that is determined by experiments. Therefore, The input series from $\bm{x}_{t-N_{\text{i}}: N_{\text{i}}}$ becomes
	\begin{equation}
	\bm{x}_{t-N_{\text{i}}: N_{\text{i}}: l}.
	\end{equation}
	The length of the input reduces from $N_{\text{p}}$ to $N_{\text{i}} =  \lceil N_{\text{p}}/l \rceil $ correspondingly, where $\lceil \cdot \rceil$ represents the operation of rounded-up. 
	\item \textbf{Model sharing.} Within each group, we train a model for all road segments, which is known as group-based model (GM). We generate the training samples for each group as 
	\begin{equation}\label{e:train}
	\bm{x}^{(r, k)} \to <\bm{x}_{t:N_{\text{i}}:l}^{(r, k)}, x_{t+N_{\text{i}}+N_{\text{o}}}^{(r, k)}>, \bm{x}^{(r, k)} \in \bm{C}^{(k)}
	\end{equation} 
	where $\bm{x}_{t:N_{\text{i}}:l}^{(r, k)}$ and $x_{t+N_{\text{i}}+N_{\text{o}}}^{(r, k)}$ denote the input and output of model, respectively. After that, we aggravate the samples within a group to train a GM $f^{(k)}(\cdot)$ for group $\bm{C}^{(k)}$,
	\begin{equation}
        x_{t+N_{\text{i}}+N_{\text{o}}}^{(r, k)} = f^{(k)}(\bm{x}_{t: N_{\text{i}}:l}^{(r, k)} ),\ \bm{x}^{(r, k)} \in \bm{C}^{(k)}
	\end{equation}  
	Then the aggregated STTP model $f(\cdot)$ at the large-scale network can be written as:
	\begin{equation}
	\begin{aligned}
        x_{t+N_{\text{i}}+N_{\text{o}}}^{(r, k)} &= f(\bm{x}^{(r, k)}_{t:N_{\text{i}}:l})\\
        &= f^{(k)}(\bm{x}_{t:N_{\text{i}}:l}^{(r, k)}),\ \  k \in \{1,2,\dots,K\},
	\end{aligned}
	\end{equation}
\end{enumerate}

\begin{figure}[t]
	\centering\includegraphics[width=3.5in]{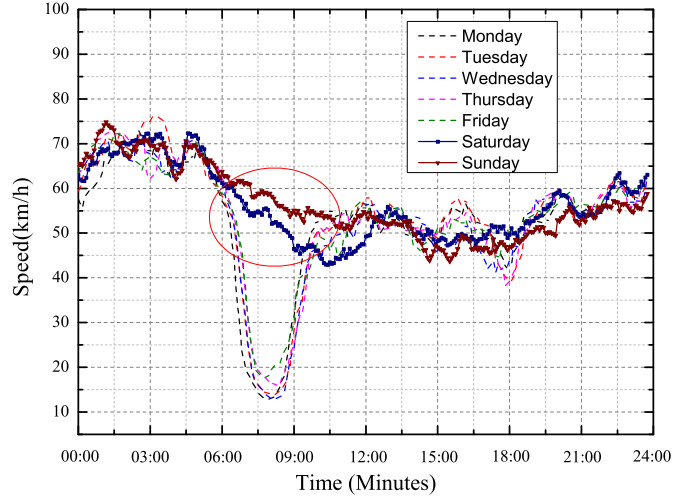}
	\caption{The every five-minute average traffic speeds of one random segment on weekdays versus to the ones at weekends from September, 2017 to November, 2017.}\label{fig:weekend}
\end{figure}

\section{Performance Evaluation}\label{sec:evaluation}
In this section, we evaluate the proposed framework on the network mentioned in Section~\ref{sec:data}. $27$ road segments are chosen for simplicity. The network, experimental settings and performance metrics are described at first. Then, we analyze the performance over different metrics.

\subsection{Experiment Settings}
\begin{figure*}[!t]
	\centering
	\subfloat[October 1]{
		\includegraphics[width=0.24\textwidth]{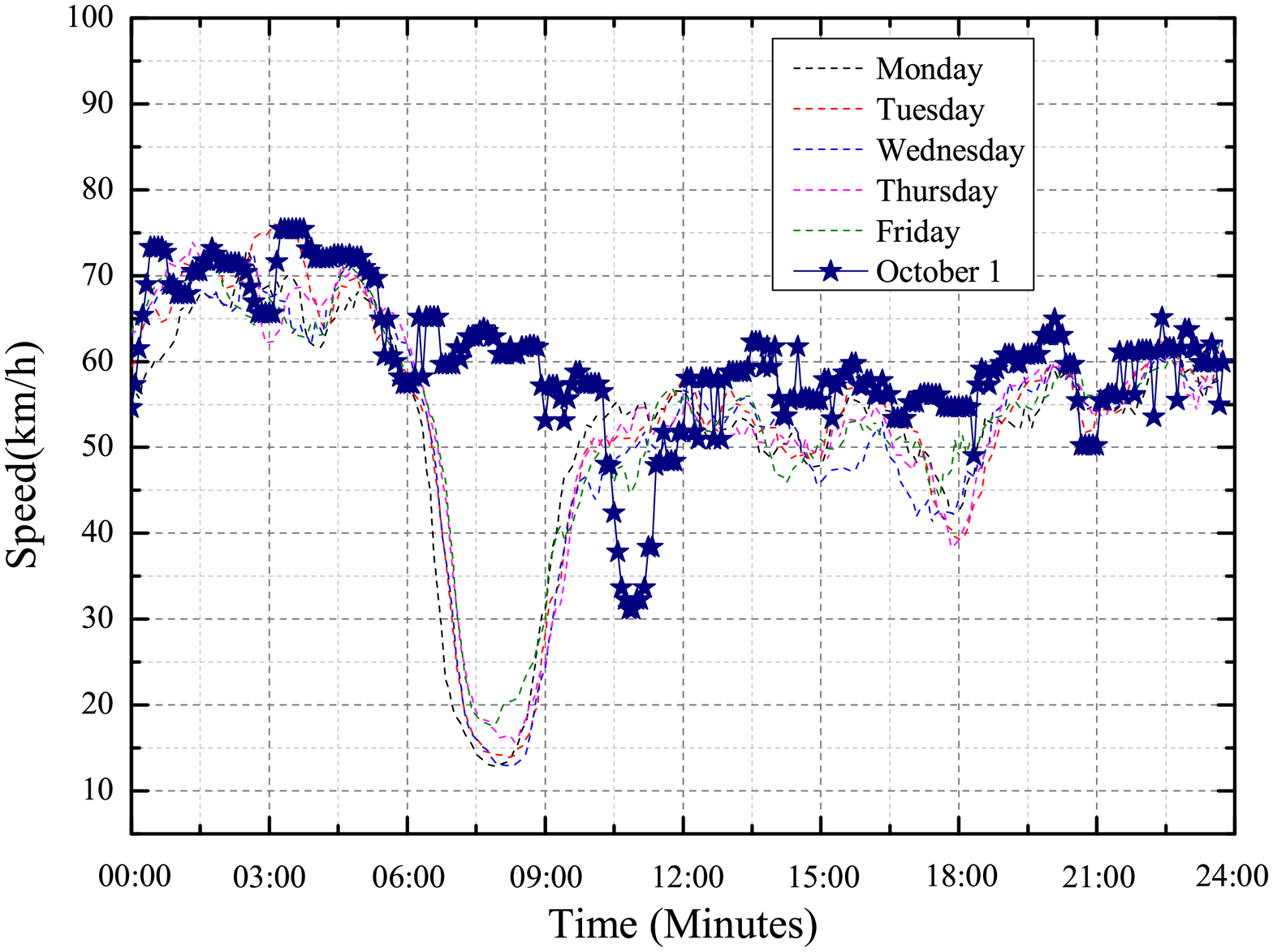}
		\label{fig:101}
	}
	\subfloat[October 2]{
		\includegraphics[width=0.24\textwidth]{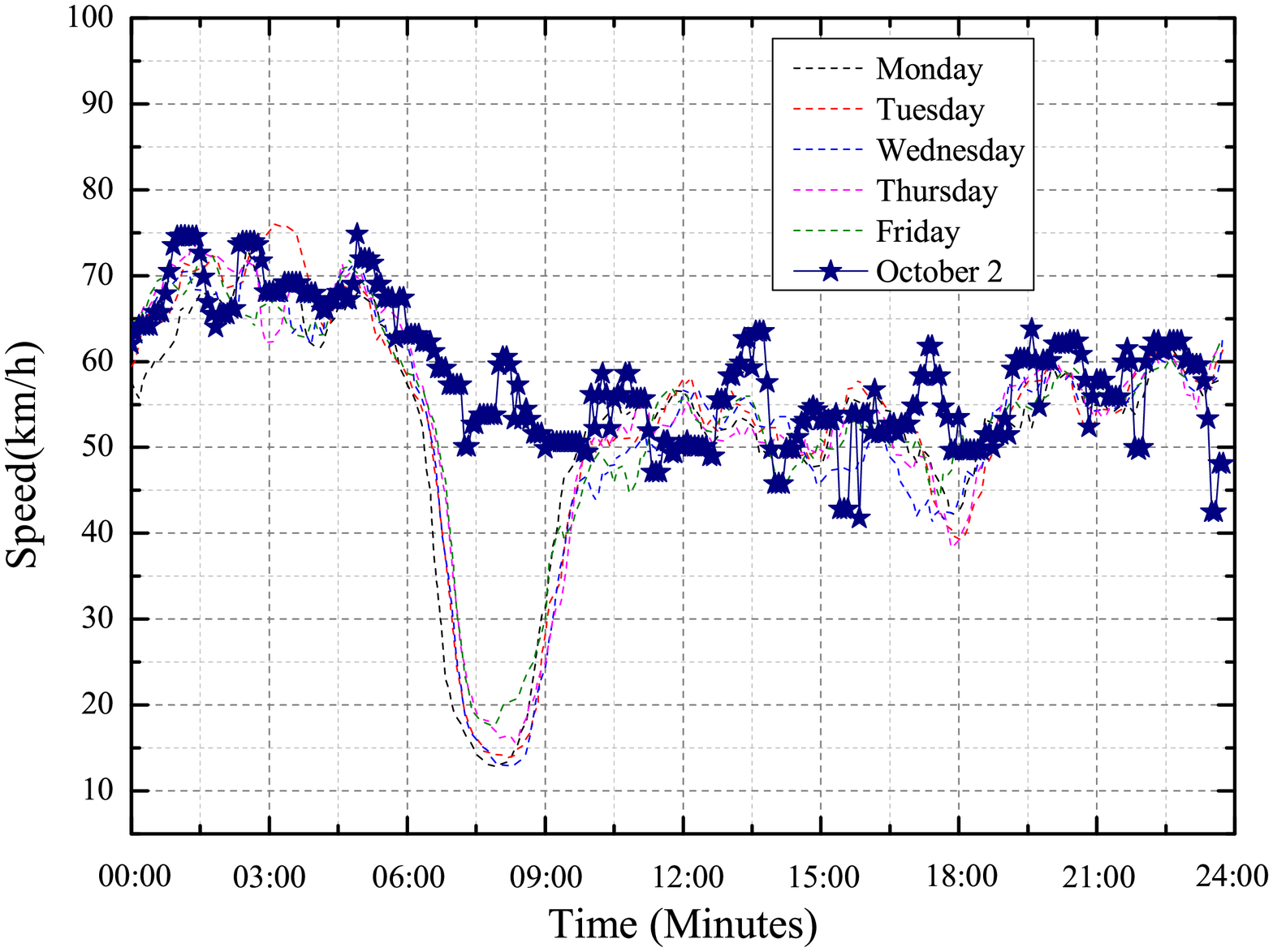}
		\label{fig:102}
	}
	\subfloat[October 3]{
		\includegraphics[width=0.24\textwidth]{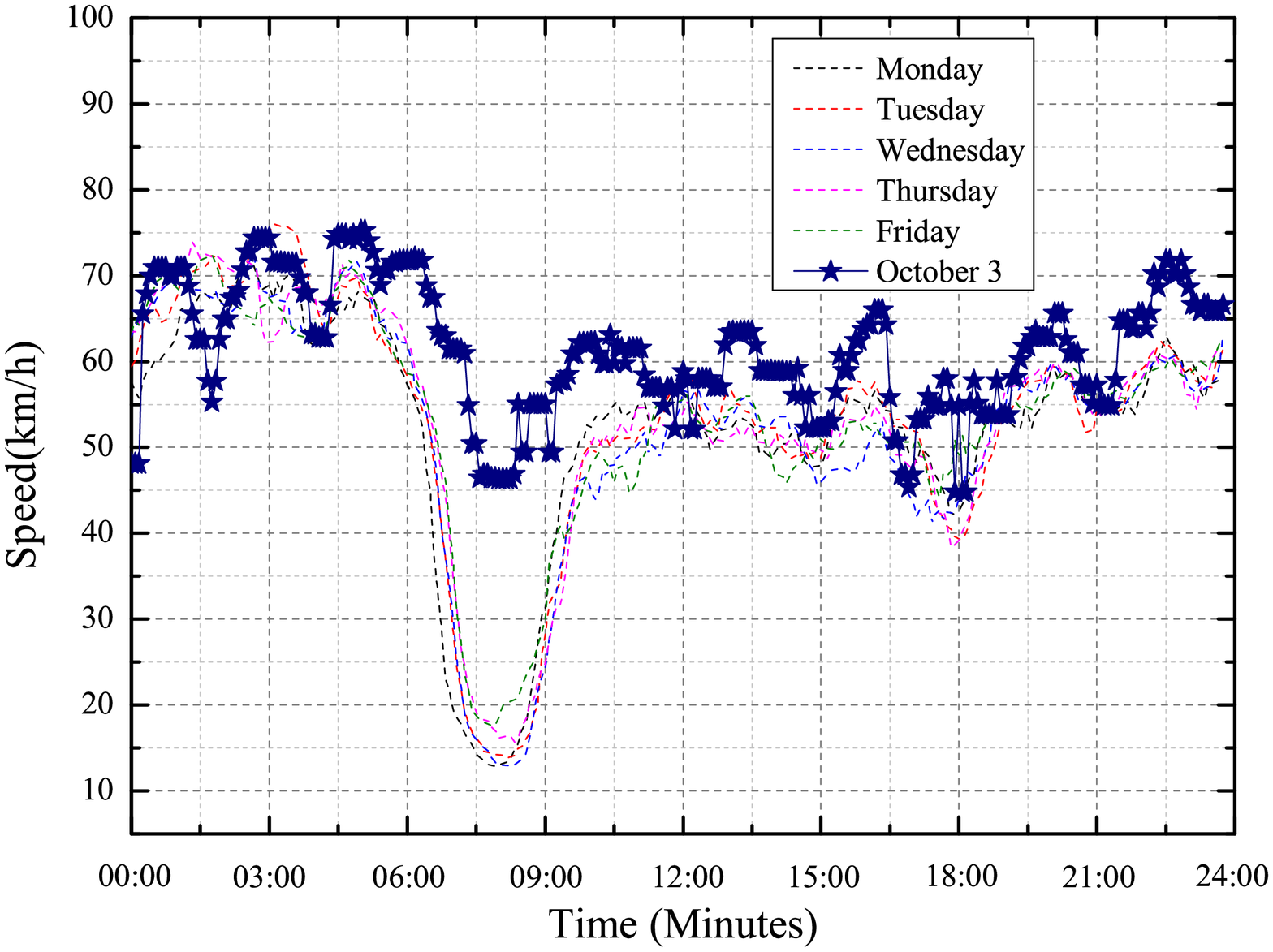}
		\label{fig:103}
	}
	\subfloat[October 4]{
		\includegraphics[width=0.24\textwidth]{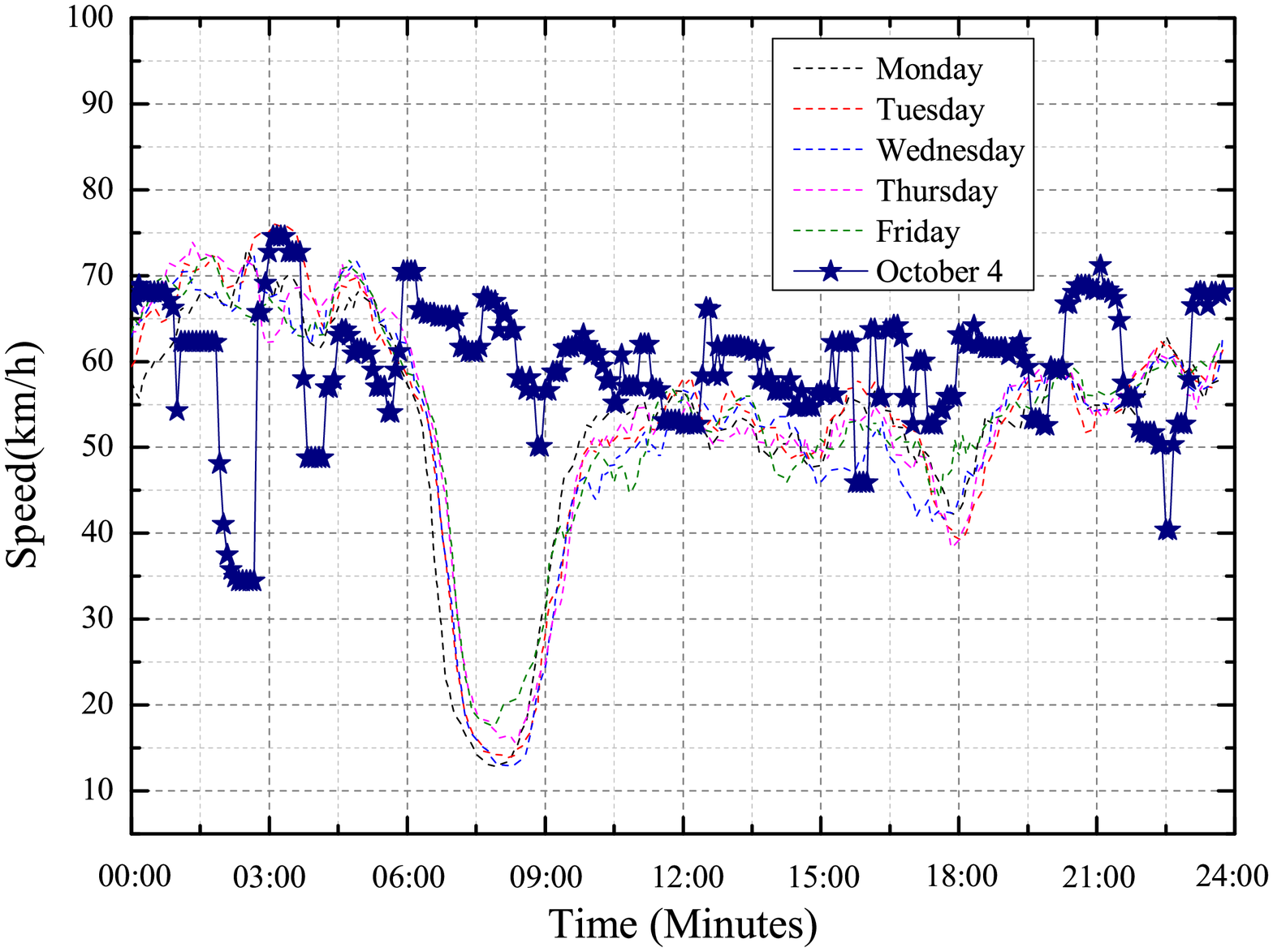}
		\label{fig:104}
	}
	\hfil
	\subfloat[October 5]{
		\includegraphics[width=0.24\textwidth]{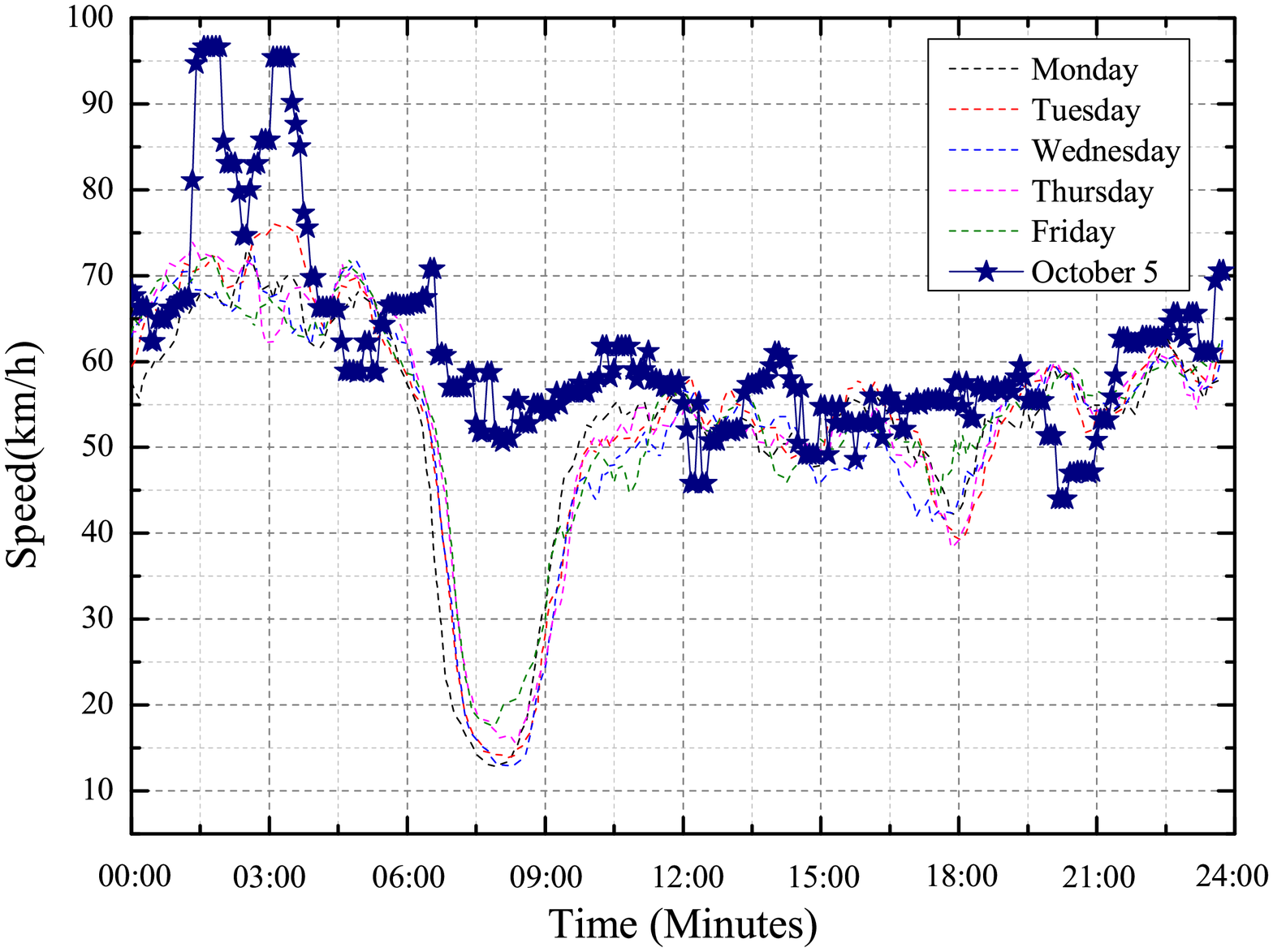}
		\label{fig:105}
	}
	\subfloat[October 6]{
		\includegraphics[width=0.24\textwidth]{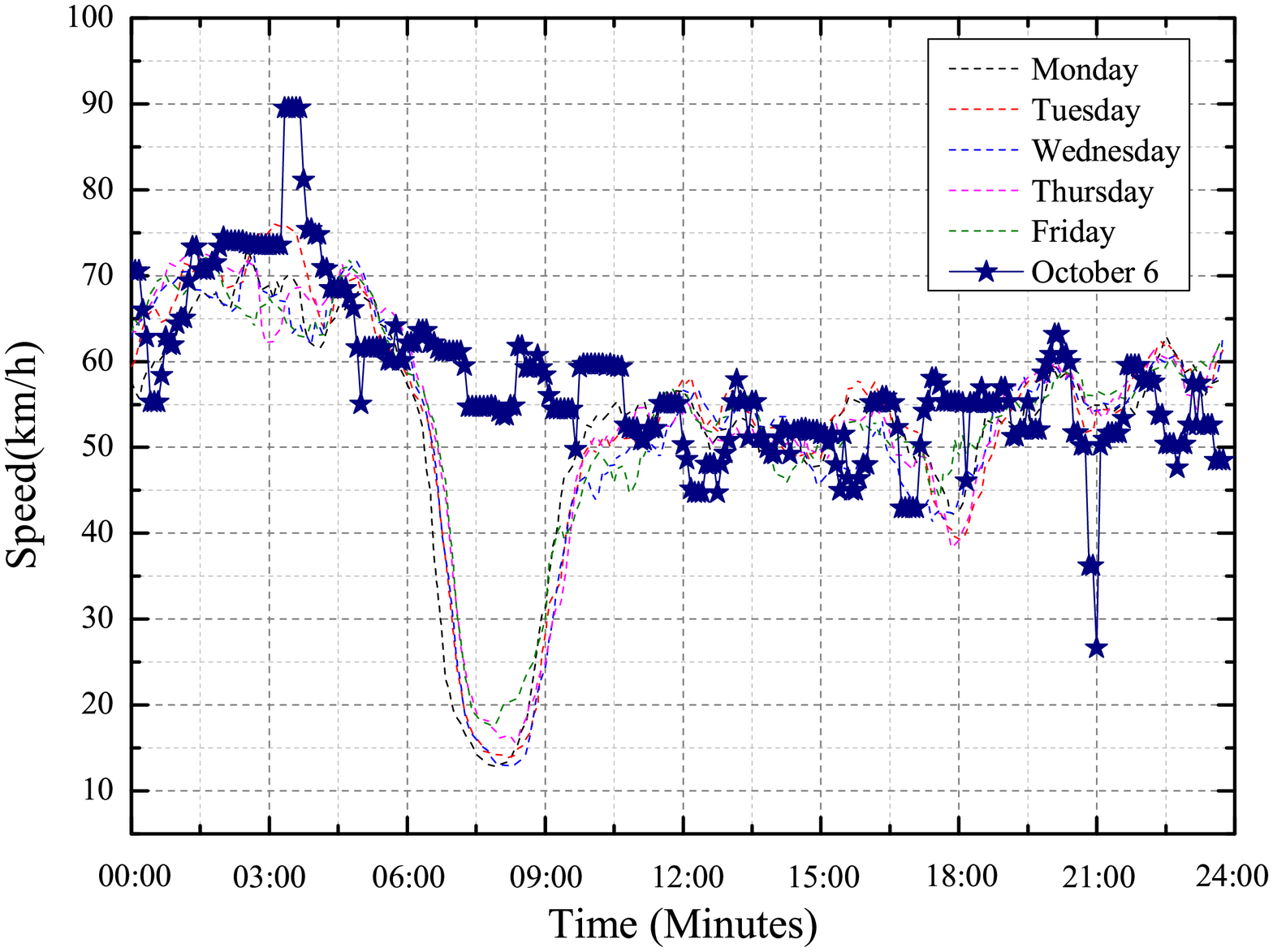}
		\label{fig:106}
	}
	\subfloat[October 7]{
		\includegraphics[width=0.24\textwidth]{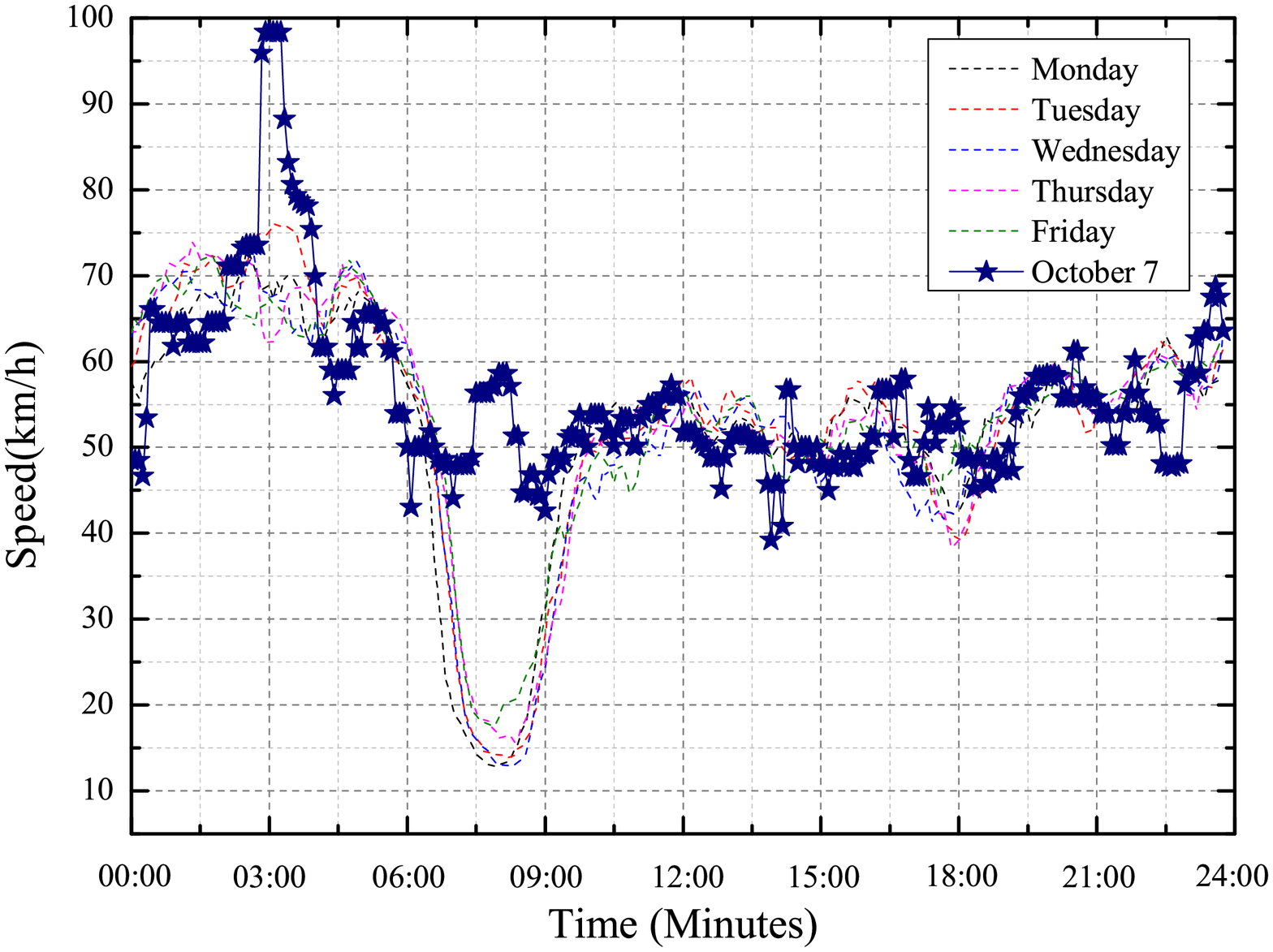}
		\label{fig:107}
	}
	\subfloat[October 8]{
		\includegraphics[width=0.24\textwidth]{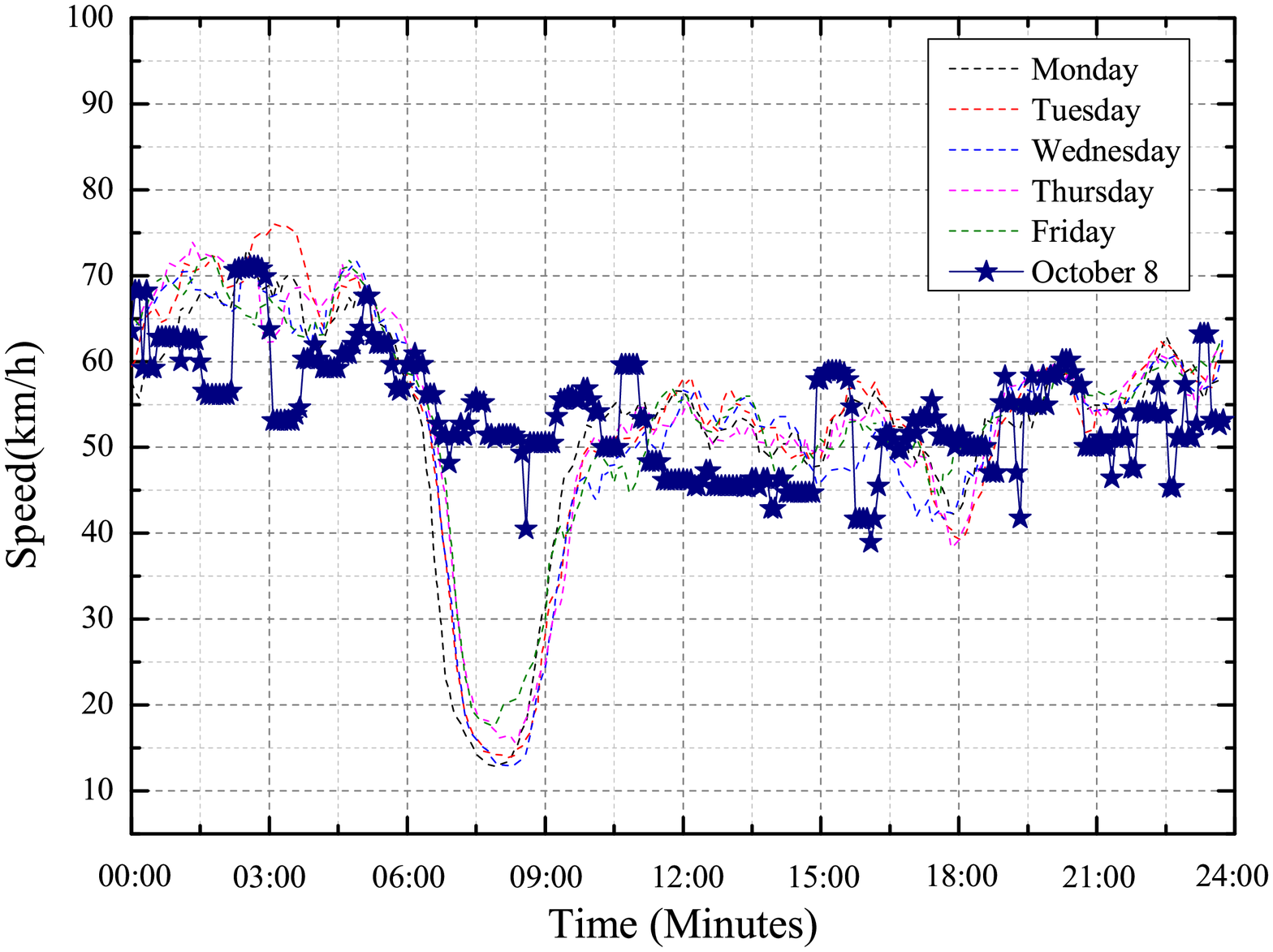}
		\label{fig:108}
	}
	\caption{The every five-minute average traffic speeds of a random segment on weekdays from September, 2017 to November, 2017 versus the ones during the National Day.}
	\label{fig:10}
\end{figure*}

For DeepCluster module, we split the traffic series into $90$ daily sub-series of length $288$ for each segment. Fig.~\ref{fig:weekend} shows that the traffic patterns on weekdays are different from the ones at weekends between six and ten o'clock in the morning, since most people do not work at weekends (The circular region). Besides, the traffic patterns behave abnormally during the National Day than usual, as shown in Fig.~\ref{fig:10}. As a result, $60$ daily sub-series are chosen by getting rid of the ones at weekends and during the National Day. Then we transfer the sub-series of size $1 \times 288$ into images of size $288\times 288$. As discussed in Section~\ref{sec:dc}, we generate triplets by the daily sub-series from $27$ road segments, which are used for representation learning. The deep structure of FaceNet used in this paper is the Inception\_ResNet, the configuration of which is the same with~\cite{facenet}. As a segment's representative, the average representations of the sub-series is used for clustering by $K$-means method. $K$ is confirmed by Silhouette coefficient~\cite{silhouette}.
\par
For DeepPrediction module, we use the state of the art RNNs, i.e., long short term memory (LSTM)~\cite{lstm} for STTP. The input span of traffic series is chosen to be a day. Then the length of the input is $ N_{\text{i}} = \lceil 288 / l \rceil $ that is confirmed by the experiments discussed later. We split the data into training set and testing set for each road segment, and aggregate the training set belonging to the same group to train the LSTM. In the end, $K$ GMs are aggregated.
\par

\begin{table}[!t]
	\label{table:network}
	\renewcommand{\arraystretch}{1.7}
	\setlength{\extrarowheight}{1pt}
	\centering
	\caption{The Configurations of the Relevant DNNs.}	
	\vspace{0.5em}
	\begin{tabular}{|c|c|c|c|}
		\hline
		\textbf{Module}&\textbf{Network} & \textbf{Parameter} & \textbf{Size} \\
		\hline
		\multirow{5}{*}{\textbf{DeepCluster}}&\multirow{6}{*}{$^a$FaceNet} & Image size & 160 \\
		\cline{3-4}
		& &Batch size& $12$ \\
		\cline{3-4}
		& &Segments per batch& $6$\\
		\cline{3-4}
		& &Images per segment& $9$\\
		\cline{3-4}
		& &Embedding size& $32$ \\
		\hline
		\multirow{5}{*}{\textbf{DeepPrediction}}&\multirow{5}{*}{$^b$LSTM}& Time steps& $\lceil 288/s \rceil$ \\
		\cline{3-4}
		& &LSTM1& $[1 \times 50]$ \\
		\cline{3-4}
		& &LSTM2& $[50 \times 25]$\\
		\cline{3-4}
		& &Dense1 & $[25\times 200] $\\
		\cline{3-4}
		& &Dense2 & $[200 \times 1] $ \\
		\hline
	\end{tabular}
	
	\vspace{2mm}
\raggedright{ {$^a$}The implications of the parameters given in this table are explained exactly in~\cite{facenet}.\\
	$^b$The inputs and outputs size are described in $[rows \times cols]$.}
\end{table}

The key parameters of the relevant DNNs are listed in Table~\ref{table:network}. If not mentioned specifically, all models are trained on eighty percent of data while tested on the remaining data. $10$-fold cross-validation is adopted over training dataset. The $K$-means method is implemented using the Scikit-learn Python 3.6.5. The NNs are conducted with a NVIDIA p2000 GPU, TensorFlow r1.8, CUDA 9.0 and CuDNN 9.0. Moreover, four performance metrics includes relative error (RE), mean relative error (MRE), max mean relative error (MARE) and minimum mean relative error (MIRE) are used for evaluation, which are defined as

\begin{equation}\label{e:re}
e_{\text{RE}}^{(r, k)} = \frac{\left| x_{t+N_{\text{o}}}^{(r, k)}-\hat{x}_{t+N_{\text{o}}}^{(r, k)} \right|}{x_{t+N_{\text{o}}}^{(r, k)}}, 1\leq k \leq K
\end{equation}  

\begin{equation}\label{e:mre}
e_{\text{MRE}}^{(k)} = \frac{1}{|\bm{C}^{(k)}|} \sum\nolimits_{\bm{x}^{(r, k)}\in \bm{C}^{(k)}} e_{\text{RE}}^{(r, k)} , 1\leq k \leq K
\end{equation}  

\begin{equation}\label{e:mare}
e_{\text{MARE}}^{(k)} = \max \limits_{\bm{x}^{(r, k)}\in \bm{C}^{(k)} } \{ e_{\text{RE}}^{(r, k)}  \}, 1\leq k \leq K
\end{equation}  

\begin{equation}\label{e:mire}
e_{\text{MIRE}}^{(k)} = \min \limits_{\bm{x}^{(r, k)}\in \bm{C}^{(k)}}\{ e_{\text{RE}}^{(r, k)}  \}, 1\leq k \leq K
\end{equation}  
where $e_{\text{RE}}^{(r, k)}$ denotes the RE of $r$-th segment in network clustered into group $\bm{C}^{(k)}$ with $x_{t+N^{\text{o}}}^{(r, k)}$ being the true speed and $\hat{x}_{t+N^{\text{o}}}^{(r, k)}$ being the prediction. $|C^{(k)}|$ is the number of road segments in the group $k$. Besides, $e_{\text{MRE}}^{(k)}$, $e_{\text{MARE}}^{(k)}$ and $e_{\text{MIRE}}^{(k)}$ are MRE, MARE and MIRE of group $k$,  respectively. The performance metrics for road network can be similarly calculated.

\begin{figure*}[!t]
	\centering
	\subfloat[Group 1]{
		\includegraphics[width=0.33\textwidth]{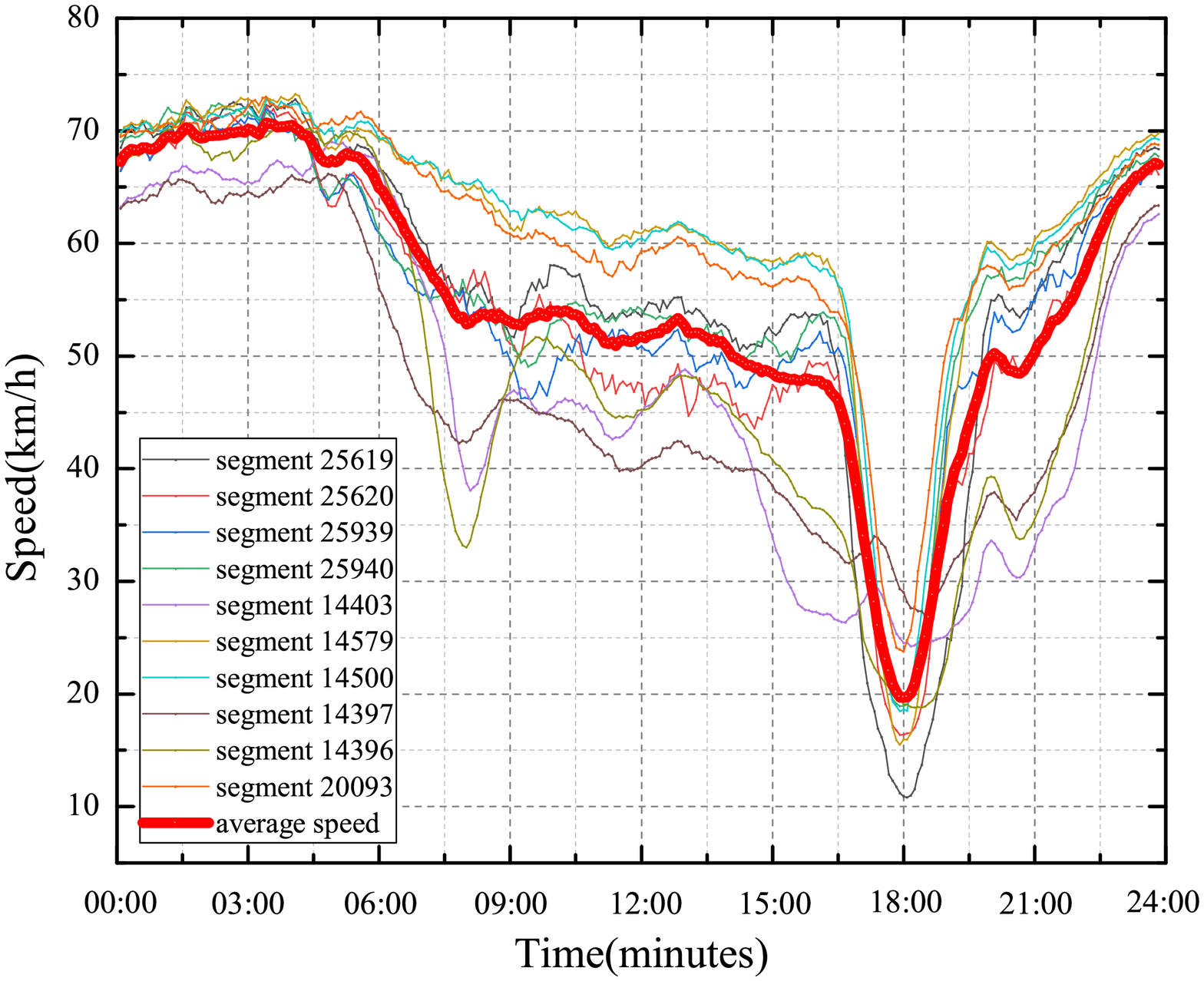}
		\label{fig:cluster1}
	}
	\subfloat[Group 2]{
		\includegraphics[width=0.33\textwidth]{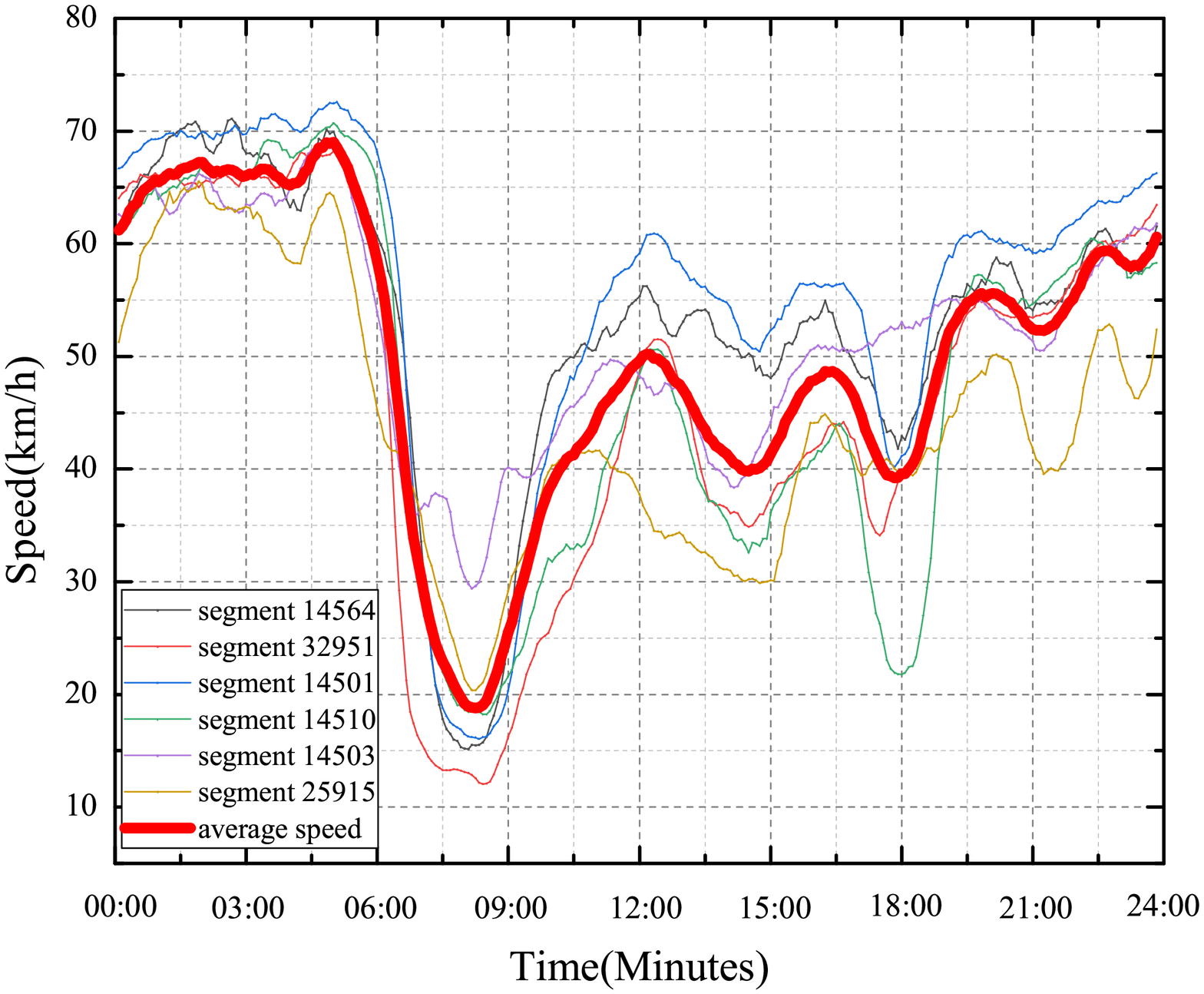}
		\label{fig:cluster2}
	}
	\subfloat[Group 3]{
		\includegraphics[width=0.33\textwidth]{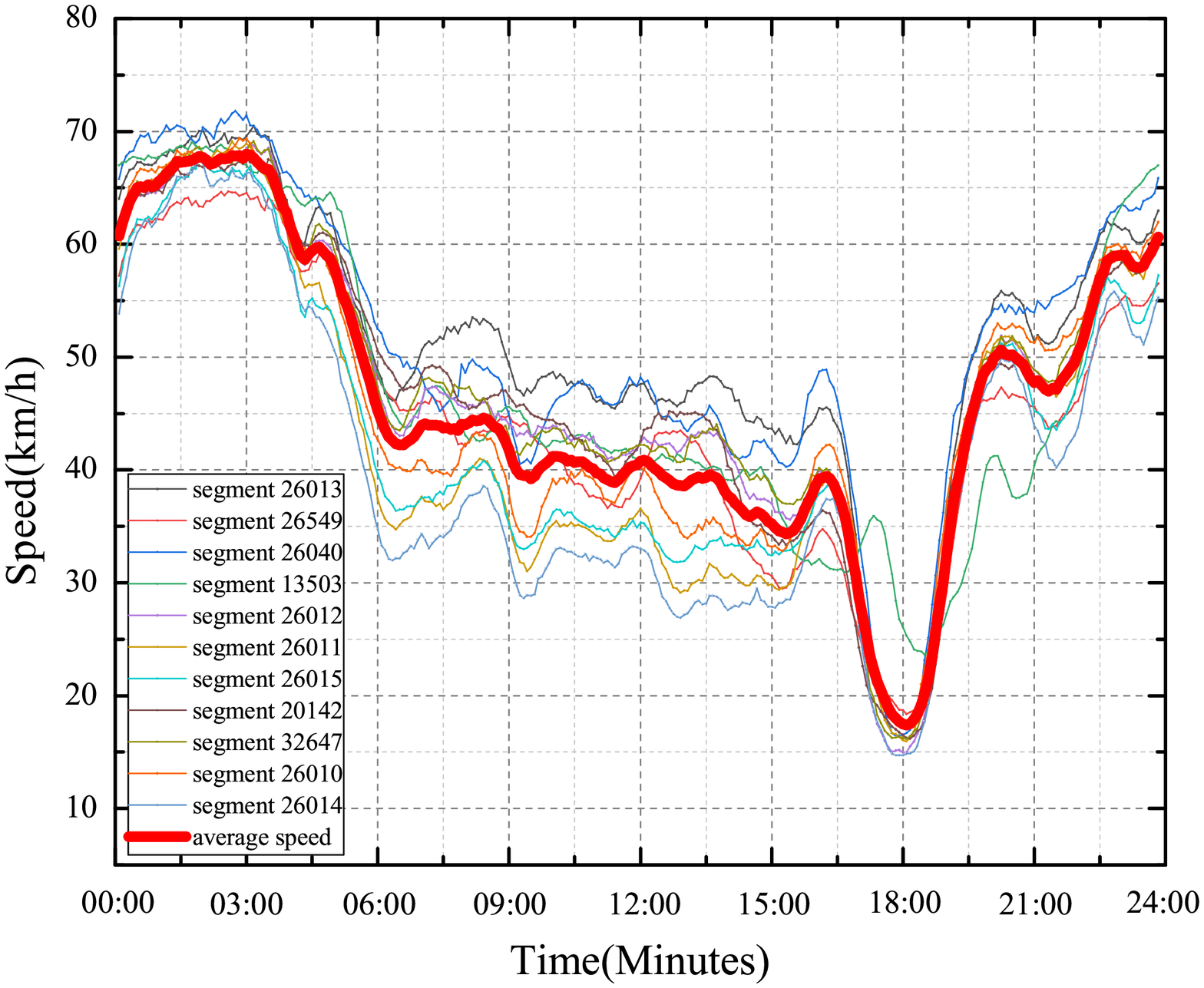}
		\label{fig:cluster3}
	}
	\caption{The every five-minute average traffic speeds of the road segments on weekdays from September, 2017 to November, 2017 in different groups. The thicker red lines represent the centers of the corresponding clusters.}
	\label{fig:cluster}
\end{figure*}

\begin{table}[!t]
	\renewcommand{\arraystretch}{1.3}
	\setlength{\extrarowheight}{1pt}
	\centering
	\caption{The Performance under Different Input Intervals}
	\label{table:interval}
	\begin{tabular}{|c|c|c|}
		\hline
		\textbf{Input interval $l$} & \textbf{MRE of Training(\%)}& \textbf{MRE of Testing(\%)}\\
		\hline		
		1& $3.70$ & $5.77$\\
		\hline
		3& $4.18$ & $5.25$\\
		\hline
		5& $4.37$ & $5.48$\\
		\hline
		7& $5.83$ & $8.77$\\
		\hline
	\end{tabular}
\end{table}

\begin{table*}[!t]
	\renewcommand{\arraystretch}{1.2}
	\setlength{\extrarowheight}{1pt}
	\centering
	\caption{The Group Performance of the Proposed Framework}
	\label{table:performance}
	\begin{tabular}{|c|c|c|c|c|c|c|c|}
		\hline
		\textbf{Prediction Horizon} & \textbf{Group} & \textbf{Algorithm}  & \makecell{\textbf{MRE of}\\ \textbf{Training($\%$)}}  & \makecell{\textbf{MRE of}\\ \textbf{Testing($\%$)}} &  \makecell{\textbf{Gap}\\ \textbf{($\%$)}} & \makecell{\textbf{MARE of}\\ \textbf{Testing($\%$)}}  & \makecell{\textbf{MIRE of}\\ \textbf{Testing($\%$)}} \\
		\hline
		\multirow{6}{*}{\textbf{1}(five-minute)} & \multirow{2}{*}{\textbf{1}}  & GM& $3.97$& $4.12$ &$0.1$ &$6.87$& $2.65$ \\
		\cline{3-8}
		& & IM & $3.20$ & $5.05$&$1.9$ &$9.80$&$3.03$\\
		\cline{2-8}		
		& \multirow{2}{*}{\textbf{2}} & GM& $4.08$ & $4.07$&$0$ & $5.76$&$3.39$\\
		\cline{3-8}
		&  & IM & $3.59$ & $4.94$&$1.3$ &$6.04$&$3.67$\\
		\cline{2-8}		
		& \multirow{2}{*}{\textbf{3}}  & GM & $4.96$ & $5.00$&$0$ & $6.54$&$4.39$\\		
		\cline{3-8}
		&  & IM& $3.72$ & $5.37$&$1.6$ &$6.86$&$4.40$\\
		\hline
		\multirow{6}{*}{\textbf{2}(Ten-minute)} & \multirow{2}{*}{\textbf{1}}  & GM& $5.92$ & $6.04$&$0.1$ & $10.07$&$3.70$ \\
		\cline{3-8}
		& & IM& $3.80$ & $5.77$&$2.0$&$9.94$&$3.57$\\
		\cline{2-8}		
		& \multirow{2}{*}{\textbf{2}} &GM & $6.22$ & $6.22$&$0$& $9.86$&$5.29$\\
		\cline{3-8}
		&  & IM & $3.90$ &$5.67$&$1.8$&$6.70$&$4.84$\\
		\cline{2-8}		
		& \multirow{2}{*}{\textbf{3}}  &  GM &$7.16$ & $7.24$&$0$&$9.56$&$6.27$\\	
		\cline{3-8}
		&  & IM & $4.20$ & $6.17$&$2.0$&$7.73$&$4.91$\\
		\hline
		\multirow{6}{*}{\textbf{3}(Fifteen-minute)} & \multirow{2}{*}{\textbf{1}} &GM& $7.08$ & $7.35$&$0.3$& $11.82$&$4.61$\\
		\cline{3-8}
		& & IM& $4.01$ & $5.82$&$1.8$&$9.21$&$3.97$\\
		\cline{2-8}		
		& \multirow{2}{*}{\textbf{2}} & GM& $7.71$ & $7.93$&$0.2$& $11.54$&$6.68$\\
		\cline{3-8}
		&  & IM& $4.12$ & $5.66$&$1.5$&$7.00$&$4.69$\\
		\cline{2-8}		
		& \multirow{2}{*}{\textbf{3}}  & GM& $8.36$ & $8.43$&$0.1$& $12.00$&$7.07$\\
		\cline{3-8}
		&  & IM& $4.71$ & $6.38$&$1.7$&$8.49$&$4.87$\\
		\hline	
	\end{tabular}

\vspace{2mm}
\raggedright{GM: Group-based Model. IM: Individual-based Model.}
\end{table*}

\subsection{Simulation Results}
\begin{figure}[t]
	\centering\includegraphics[width=3.5in]{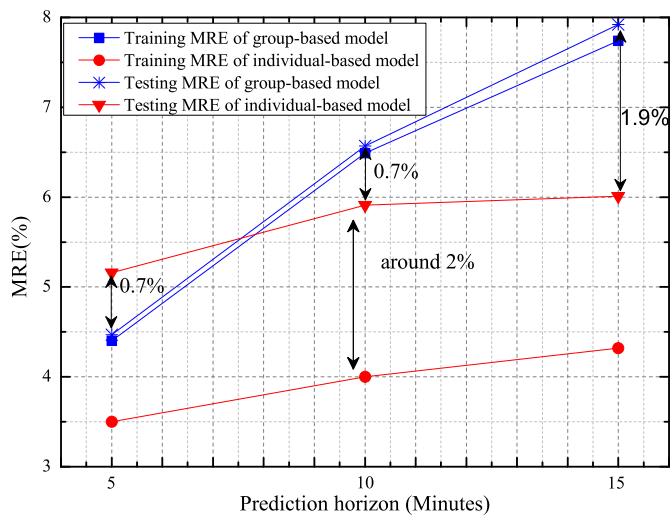}
	\caption{The network MRE of GM and IM.}\label{fig:network3}
\end{figure}

\begin{figure}[t]
	\centering\includegraphics[width=3.5in]{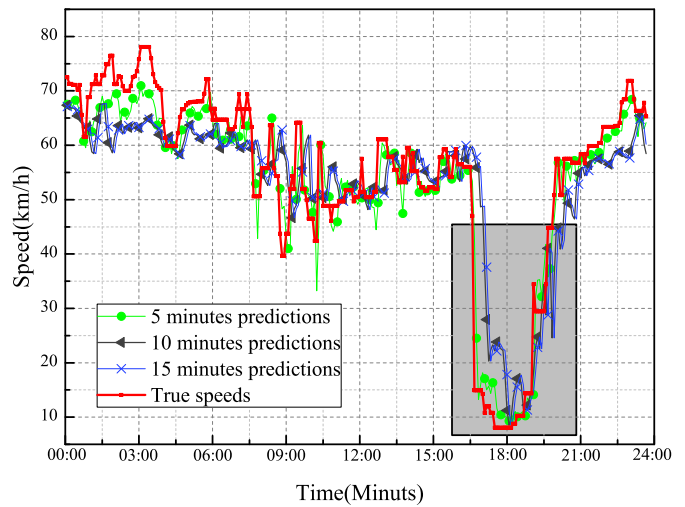}
	\caption{The every five-minute predictions using GM model versus true speeds of a random road segment in group 1 on November 31, 2017.}\label{fig:vivid}
\end{figure}

Three experiments are conducted, including road segments clustering, interval confirmation and STTP at network.
\begin{enumerate}	
\item \textbf{Road segments clustering.} All 27 road segments are clustered into $3$ groups by DeepCluster as shown from Fig.~\ref{fig:cluster1} to Fig.~\ref{fig:cluster3}. It can be found that the series in a group are in general \textit{homogeneous} with the other series defined at Section~\ref{sec:dc}, which demonstrates the proposed DeepCluster's capacity of extracting the shape-based features. For example, the segments in cluster 1 have a breakdown in traffic speed during the evening peak period, followed by speed recovery. The cluster 2 have a breakdown during the morning peak, and start to swing at the middle speed back-and-forth. The segments in cluster 3 have some slight resemblances to cluster 1 during the evening peak period. However there is a stable condition holding the middle speed after six o 'clock in the morning.
\par	
\item \textbf{Interval confirmation.} This part investigates the effect of input interval on predictive performance and determines the threshold $p$ of the ACF defined in Section~\ref{sec:dp}. The LSTM is performed to predict the next five-minute speed under different input intervals $l$ over the $3$ random segments. From the performance listed in Table~\ref{table:interval}, the MRE of training increases with the decrease of $l$. However, the performance improvements are insignificant when $l\leq 5$, such as the training MRE at $3.7\%$ and $4.4\%$ when $l=1$ and $l=5$. Besides, the testing MRE at $l=1$ is slightly larger than that at $l=5$. This is because the capacity of model becomes stronger as input interval decreases, leading to overfitting. From this result, the threshold is empirically set to $0.8$. In the end, the input interval is set to $5$ corresponding to twenty-five minutes for all other simulations.
\par
\item \textbf{STTP at network.} For the performance comparison, we construct an IM for a segment by the same configuration of LSTM under different prediction horizon $N_{o}$. Simulation results are listed in Table~\ref{table:performance}. The IMs have lower training MRE than the GMs due to the fact that the capacity of the IMs is highly stronger than that of the GMs. However, the GMs can get lower gaps between training MRE and testing MRE in all tests, since increasing the number and diversity of the training samples can improve generalization capability of the model. On the contrary, the IMs are constrained by the problem of overfitting resulted from modeling the noise. As shown in Fig.~\ref{fig:network3}, the gaps of GMs are close to $0$ while the gaps of IMs are around $2\%$. 
\par
From Table~\ref{table:performance}, we can observe that the GMs perform better than IMs in terms of testing error in a relatively simple task of five-minute forecasting. The testing MRE of the GMs and IMs are $4.12\%$ and $5.05\%$ for group 1, $4.07\%$ and $4.94\%$ for group 2, $5.00\%$ and $5.37\%$ for group 3, respectively. However, as the task becomes complex, the capacity of GMs become insufficient. For example, the testing MRE of GMs are around $1\%$ more than that of IMs when $N_{\text{o}}=2$, while the testing MRE of GMs are around $2\%$ more than that of IMs when $N_{\text{o}}=3$. 
\par
As shown in Fig.~\ref{fig:vivid}, the GM can predict the trends of traffic speed well, but the performance gets worse with the increase of the prediction horizon. It also shows that the model does not work well of 10 and 15 minutes forecasting in rush hours (The dash area in Fig.~\ref{fig:vivid}), that the traffic speed switching sharply. 
\par
The proposed framework is scalable that can be applied for the large-scale networks easily by reducing the number of models significantly, and can reach the compromise of the number of models and prediction performance. Compared to the traditional $27$ IMs, the number of prediction models has been reduced up to $\frac{(27-3)}{27} \approx 88\%$ with about $0.7\%-1.9\%$ performance degradation, in terms of network MRE in our test, as shown in Fig.~\ref{fig:network3}. In conclusion, the performance of the framework is comparable to that of customized IMs, which validates the ability for STTP at large-scale networks.
\end{enumerate}

\section{Conclusion}\label{sec:conclusion}
The characteristics of the multiplicity and heterogeneity make STTP at large-scale network a challenging and important problem. By exploiting the characteristic of traffic patterns, a DL framework for STTP at large-scale networks is proposed in this paper. The key point of the framework is the combination of the DeepCluster and the DeepPrediction, as well as the model sharing strategy. We analytically evaluate the proposed framework over a real large-scale network of Liuli Bridge in Beijing and some insights into generic DL models are obtained. Despite that the prediction performances of the GMs are slightly worse than that of IMs in most tests, the GMs have a better generalization ability. For five-minute prediction, the GM gets $0.7\%$ error lower than IM. We also discuss the effect of input interval on the prediction performance, which guides the framework on how to select the effective input interval. Furthermore, we use only $3$ models to achieve the STTP at network, while the traditional way needs to construct $27$ models. 

\section{Acknowledgement}
This work is funded in part by the National Natural Science Foundation of China under Grant 61731004.

\vskip 0pt plus -1fil
\begin{IEEEbiographynophoto}
{Lingyi Han} received the B.S. degree from Beijing University of Posts and Telecommunications, China, in 2016. She is currently pursuing the Ph.D. degree with the Intelligent Computing and Communication Laboratory, Key Laboratory of Universal Wireless Communications, Ministry of Education, Beijing University of Posts and Telecommunications. His research interests include data mining and artificial intelligence in Internet-of-Things.
\end{IEEEbiographynophoto}
\vskip 0pt plus -1fil
\begin{IEEEbiographynophoto}
{Kan Zheng} received the B.S., M.S., and Ph.D. degrees from the Beijing University of Posts and Telecommunications (BUPT),  Beijing, China, in 1996, 2000, and 2005, respectively. He is currently a Professor at BUPT. He is the author of more than 200 journal articles and conference papers in the field of resource optimization in wireless networks, M2M networks, VANET, and so on. He has rich industry experiences on the standardization of the new emerging technologies. Dr. Zheng holds editorial board positions for several journals. He has organized several special issues in famous journals, including the IEEE COMMUNICATIONS ON SURVEYS AND TUTORIALS and TRANSACTIONS ON EMERGING TELECOMMUNICATIONS TECHNOLOGIES. 
	\end{IEEEbiographynophoto}
\vskip 0pt plus -1fil
\begin{IEEEbiographynophoto}
{Long Zhao} received the B.S. degree from Nanjing University of Posts and Telecommunications, Nanjing, China, in 2008 and the M.S. degree from Harbin Engineering University, Harbin, China, in 2011. He is currently working toward the Ph.D. degree with the Beijing University of Posts and Telecommunications, Beijing. From April 2014 to March 2015, he was a Visiting Scholar at the Department of Electrical Engineering, Columbia University, supervised by Prof. X. Wang. His research interests include wireless communications and signal processing.
\end{IEEEbiographynophoto}
\vskip 0pt plus -1fil
\begin{IEEEbiographynophoto}
{Xianbin Wang}(S'98-M'99-SM'06-F'17) received the Ph.D. degree in electrical and computer engineering from the National University of Singapore in 2001. 

From January 2001 to July 2002, he was a System Designer with STMicroelectronics, where he was responsible for the system design of DSL and Gigabit Ethernet chipsets. He was with the Communications Research Centre Canada (CRC) as a Research Scientist/Senior Research Scientist from 2002 to 2007. He is currently a Professor and the Canada Research Chair with Western University, Canada. His current research interests include 5G technologies, Internet-of-Things, communications security, and locationing technologies. He has over 300 peer-reviewed journal and conference papers, in addition to 26 granted and pending patents, and several standard contributions.

Dr. Wang is an IEEE Distinguished Lecturer. He received many awards and recognition, including the Canada Research Chair, the CRC Presidents Excellence Award, the Canadian Federal Government Public Service Award, the Ontario Early Researcher Award, and five IEEE Best Paper Awards. He was involved in a number of IEEE conferences, including GLOBECOM, ICC, VTC, PIMRC, WCNC, and CWIT, in different roles, such as the Symposium Chair, a Tutorial Instructor, the Track Chair, the Session Chair, and the TPC Co-Chair. He currently serves as an Editor/Associate Editor of the IEEE TRANSACTIONS ON COMMUNICATIONS, the IEEE TRANSACTIONS ON BROADCASTING, and the IEEE TRANSACTIONS ON  VEHICULAR TECHNOLOGY. He was also an Associate Editor of the IEEE TRANSACTIONS ON WIRELESS COMMUNICATIONS from 2007 to 2011, and the IEEE WIRELESS COMMUNICATION LETTERS from 2011 to 2016.
\end{IEEEbiographynophoto}
\vskip 0pt plus -1fil
\begin{IEEEbiographynophoto}
{Xuemin (Sherman) Shen} (M'97-SM'02-F'09) received the B.Sc. degree from Dalian Maritime University, Dalian, China, in 1982 and the M.Sc. and Ph.D. degrees from Rutgers University, Piscataway, NJ, USA, in 1987 and 1990, all in electrical engineering.

From 2004 to 2008, he was the Associate Chair for Graduate Studies with the Department of Electrical and Computer Engineering, University of Waterloo, Waterloo, ON, Canada. He is currently a Professor and the University Research Chair with the Department of Electrical and Computer Engineering, University of Waterloo. He is a coauthor or editor of six books and the author of several papers and book chapters in wireless communications and networks, control, and filtering. His research interests include resource management in interconnected wireless/ wired networks, wireless network security, wireless body area networks, and vehicular ad hoc and sensor networks.

Dr. Shen served as the Technical Program Committee Chair for the 2010 Fall IEEE Vehicular Technology Conference (IEEE VTC’10 Fall); the Symposia Chair for the 2010 IEEE International Conference on Communications (IEEE ICC’10); the Tutorial Chair for IEEE VTC’11 Spring and IEEE ICC’08; the Technical Program Committee Chair for the 2007 IEEE Global Communications Conference; the General Co-Chair for the 2007 IEEE International Conference on Communications and Networking in China and the 2006 Third International Conference on Quality of Service in Heterogeneous Wired/ Wireless Networks; and the Chair for IEEE Communications Society Technical Committee on Wireless Communications and Peer-to-Peer Communications and Networking. He also serves/served as the Editor-in-Chief for IEEE NETWORK, Peer-to-Peer Networking and Application, and IET Communications; as a Founding Area Editor for IEEE TRANSACTIONS ON WIRELESS COMMUNICATIONS; as an Associate Editor for IEEE TRANSACTIONS ON VEHICULAR TECHNOLOGY, Computer Networks, and ACM Wireless Networks; and as the Guest Editor for IEEE JOURNAL ON SELECTED AREAS IN COMMUNICATIONS, IEEE WIRELESS COMMUNICATIONS, IEEE COMMUNICATIONS MAGAZINE, and ACM Mobile Networks and Applications. He is a registered Professional Engineer of Ontario, Canada; a Fellow of the Canadian Academy of Engineering and the Engineering Institute of Canada; and a Distinguished Lecturer of the IEEE Vehicular Technology and Communications Societies.
\end{IEEEbiographynophoto}

\end{document}